\documentclass[11pt]{article}

\usepackage[preprint]{acl}

\usepackage{times}
\usepackage{latexsym}

\usepackage[T1]{fontenc}

\usepackage[utf8]{inputenc}

\usepackage{microtype}

\usepackage{inconsolata}

\usepackage{graphicx}

\usepackage{xcolor}  
\usepackage{pxrubrica}        
\usepackage{booktabs}
\usepackage{multirow}
\usepackage{xspace}
\usepackage{amsmath}
\usepackage{amssymb}
\usepackage{subfigure}
\usepackage{comment}
\usepackage{makecell}
\usepackage{color}
\usepackage{pifont}
\usepackage{tabularx}
\usepackage{colortbl}
\usepackage{listings}
\usepackage{caption}
\usepackage{enumitem}
\usepackage{courier}
\usepackage{adjustbox}
\usepackage{array}

\definecolor{red}{rgb}{0.8,0,0}
\definecolor{green}{rgb}{0,0.8,0}
\definecolor{blue}{rgb}{0,0,0.8}
\definecolor{yellow}{rgb}{0.90,0.91,0.75}
\definecolor{purple}{rgb}{0.85,0.80,0.91}

\definecolor{codegreen}{rgb}{0,0.6,0}
\definecolor{codegray}{rgb}{0.5,0.5,0.5}
\definecolor{codepurple}{rgb}{0.58,0,0.82}
\definecolor{codeblue}{rgb}{0.25,0.5,0.75}
\definecolor{backcolour}{rgb}{0.95,0.95,0.92}
\definecolor{codeorange}{rgb}{1,0.6,0}
\definecolor{codeyellow}{rgb}{0.7,0.7,0}
\definecolor{codeteal}{rgb}{0,0.5,0.5}

\title{Neuron Level Analysis of Large Language Model \\ in Legal Domain Reasoning}

\author{
    Eri Onami$^{1,2}$   
    Youmi Ma$^{1}$ 
    Shuhei Kurita$^{2,1}$ 
    Naoaki Okazaki$^{1,2,3}$\\
      $^{1}$ Institute of Science Tokyo 
      $^{2}$ NII 
      $^{3}$ AIST  \\
      \texttt{eri.onami at nlp.comp.isct.ac.jp},
      \texttt{ ma.y at comp.isct.ac.jp},\\
      \texttt{ skurita at nii.ac.jp},
      \texttt{ okazaki at comp.isct.ac.jp} \\}

\begin{document}
\maketitle

\begin{abstract}

We presented a neuron-level analysis of legal-domain reasoning in LLMs, comparing it with other applied domain tasks across seven open-weight models.
Using neuron attribution scores to rank and suppress influential neurons, we confirmed that suppressing the identified neurons collapses accuracy on the target task, whereas suppressing the same number of random neurons does not.
We further found a small subset of neurons influential across all seven tasks; once these are removed, suppressing the remaining neurons degrades only the task they were identified from, revealing genuinely task-specific neurons in every model studied.
Within the legal domain, the three benchmarks exhibit relatively high neuron overlap and tend to be affected jointly, suggesting of legal components neurons that span jurisdictions.
The distribution of identified neurons in our experiments suggests that the hypothesis that influential neurons are concentrated in middle MLP layers may depend on the input format and content, rather than being a universal phenomenon.

\end{abstract}

\section{Introduction}

With the rapid advancement of large language models (LLMs), general tasks such as everyday conversation, translation, and coding have reached a practically usable level.
In contrast, performance in applied domains such as law, mathematics, and medicine remains insufficient~\cite{cobbe2021gsm8k,jin2021disease,fan2025lexam,lexeval2024}.
In the legal domain, general open-source LLMs have not yet reached a passing level on multiple-choice questions of bar examinations~\cite{cao2025jbeqajapanesebarexam}, indicating the need for additional adaptation.
Although the development of sufficiently large datasets is essential for improving the performance of LLMs, it is difficult to construct complete datasets for application domain tasks because of information protection, personal privacy protection, and the protection of trade secrets in legal domain. 
Also, the model may face unprecedented legal cases in which both parties to the dispute have rights worthy of protection.
As a result, it is not realistic to develop datasets covering all categories of problems that LLMs are expected to solve.
Furthermore, a substantial share of high-quality LLM training data is concentrated in English, whereas legal systems differ fundamentally across jurisdictions. 
Knowledge learned from one jurisdiction is often not transferable to another, raising serious concerns about degraded performance in non-English legal systems, where models may fail to capture jurisdiction-specific legal rules and reasoning structures.

Previous studies analyzed the internal behavior
of transformer-based pretrained language model. Understanding of LLM's internal components has been proved to be useful for increasing model abilities such as bias mitigation with activated neurons~\cite{yang-etal-2024-mitigating} and long context understanding with using attention heads~\cite{ma2026interpretabilityperformanceoptimizingretrieval}.
LLM interpretability studies on the internal mechanisms of language models contribute to the understanding of models' behaviors without relying on extensive computational approaches such as additional training.
Several prior works have identified language neurons associated with specific model ability, such as social bias~\cite{yang-etal-2024-mitigating}, query-relevant neurons~\cite{query_relavant_neuron}, language-specific neurons~\cite{tang-etal-2024-language,kojima-etal-2024-multilingual}, and culture neurons~\cite{ying-etal-2025-disentangling,namazifard-poech-2025-isolating,yamamoto2026neuronlevel}.
If neurons associated with the legal domain can be identified, there is potential to improve task performance without additional model training because research on LLM interpretability aims not only to clarify the internal mechanisms of LLMs but also to leverage such insights for performance improvement.

In this paper, we investigate whether legal-domain-specific neurons do exist, analogous to task-specific neurons in other domains, by comparing them with neurons that are specific to other applied-domain tasks, such as mathematics, medicine, common sense reasoning, and translations.
Our research questions are: (i) are these task-specific neurons are unique to their tasks or widely shared across tasks, (ii) do the legal domain specific neurons that contribute to a wide range of legal tasks exist, (iii) the relationships and differences between legal task specific neurons and other task specific neurons in LLMs, and (iv) the distribution of those task specific neurons in the large language models across layers, attention and MLP components.
To address these questions, we analyze seven domain application tasks from legal of LexEval~\cite{lexeval2024}, LexGLUE~\cite{chalkidis-etal-2022-lexglue}, and LEXam~\cite{fan2025lexam}, mathematics of GSM8K~\cite{cobbe2021gsm8k}, medicine of MedQA~\cite{jin2021disease}, physical/spatial common sense reasoning of PIQA~\cite{Bisk2020}, and translation of WMT19~\cite{barrault-etal-2019-findings} to capture each tasks specific neurons and examine whether the observed patterns are specific to the legal domain or instead generalize across applied reasoning tasks.

In experiments, suppressing task-specific neurons significantly degrades the accuracy of that task, while we notice that this task-specific neuron suppression doesn't largely affect the accuracy of another task, especially for tasks that are in distant backgrounds, e.g., legal tasks and others.
Furthermore, we also notice that there are a small portion of neurons that are influential to multiple tasks across domains. 
When we remove such across-task influential neurons from task-specific neurons, this strengthens the task-uniqueness of the remaining neurons.
In legal domain-specific neuron analysis, we observe two main phenomena: suppressing identified legal task neurons in one task decline the other legal task accuracy, and legal neurons have relatively larger overlap with each other.
In discussion, the distribution of identified neurons in our experiment shows that the hypothesis of influential neuron distribution in middle MLP layers may have input-text format and content dependency rather than a universal phenomenon.

\section{Related Work}
\label{sec:related}

\subsection{LLM interpretability}
Mechanistic interpretability~\cite{nanda2023progress} research has aimed to reveal how information is represented within individual LLM components including attention heads~\cite{crosbie-shutova-2025-induction, wu2025retrieval}, projecting model internal vectors to vocabulary space~\cite{dar-etal-2023-analyzing,stolfo2023a}, and neuron level analysis.
Prior studies have identified individual neurons and neuron populations associated with particular capabilities, such as knowledge storage neurons~\cite{dai-etal-2022-knowledge}, linguistic neurons that are activated by LLM inputs of specific languages~\cite{kojima-etal-2024-multilingual, tang-etal-2024-language}, and domain specific neurons in vision language models~\cite{huo-etal-2024-mmneuron}.
However, these studies often lack to examine whether their methods and identified neurons are task-specific by comparing with controlled experiments on other domains.
Despite the growing attention on improving LLM reasoning abilities, there has been very little research analyzing LLM interpretability in reasoning tasks outside of mathematics.

\subsection{LLM in Legal Domain}
\label{subsec:rw_legal}

Legal question-answering datasets requiring legal reasoning ability have been proposed across multiple jurisdictions and languages.
LEXam~\cite{fan2025lexam} collects university-level examination QA in Switzerland in both multiple-choice and open-ended formats.
LexEval~\cite{lexeval2024} provides a comprehensive Chinese legal benchmark covering 23 tasks and over 14k questions.
LexGLUE~\cite{chalkidis-etal-2022-lexglue} collects QA datasets in several languages, including US law.
Existing methods to improve LLM legal performance largely rely on continued pretraining or supervised finetuning on jurisdiction-specific corpora.
However, for the reason of personal privacy,  company's trade secrets, and intellectual property rights protection, it is not feasible to construct datasets that comprehensively cover every type of problem LLMs may encounter, research is needed on improving legal generation capabilities through approaches other than additional training.  

\subsection{LLM in Application Domain}
In applied domains such as mathematics~\cite{lightman2023lets}, medicine~\cite{sun-etal-2025-reasonmed, zhang-etal-2025-generation, sviridova-etal-2024-casimedicos},  and finance~\cite{reddy-etal-2024-docfinqa, magomere-etal-2025-finnli}, a wide range of benchmark datasets have been proposed and are being used in combination with diverse training methods, including DPO~\cite{rafailov2023direct} and GRPO~\cite{deepseek-math}.
Particularly in mathematics, research has examined the internal mechanisms of LLMs using methods such as causal analysis to investigate mathematical reasoning~\cite{nikankin2025arithmetic}, the mechanisms of reasoning steps, and that a large proportion of reasoning steps are not in fact essential to the reasoning process itself~\cite{nikankin2026reasoningmodelsknowwhats}.

\section{Method}

We aim to identify neurons that are indispensable for reasoning, in the sense that suppressing their outputs leads to a substantial drop in the model's overall reasoning accuracy.
Our proposal is built on neuron-level attribution methods from the mechanistic interpretability literature, which assign an importance score to each neuron with respect to a given task.
We first identify a small set of neurons that are most influential for solving a target task using a training split, and we then \emph{suppress} their activations at inference time on the test split.

\subsection{Background: Neuron Attribution Score}
\label{subsec:neuron_attribution_background}

A line of work in mechanistic interpretability has formalized the question of which neurons are responsible for a particular model behavior.
\citet{dhamdhere2018important} extended Integrated Gradients~\cite{sundararajan2017axiomatic} from input attribution to internal neurons via the \emph{conductance}, which integrates the gradient of the output with respect to a neuron's activation along a straight-line path from a reference activation to the actual one.
\citet{dai-etal-2022-knowledge} adapted conductance to the FFN sub-layers of Transformer language models and used it to localize \emph{knowledge neurons} that store factual associations, validating their causal role by suppressing or amplifying their activations.
This template---compute an attribution score, select the top-$k$ neurons, and intervene to verify causal effects---has since been applied to bias neurons~\cite{yang-etal-2024-mitigating}, query-relevant neurons~\cite{query_relavant_neuron}, language-specific neurons~\cite{tang-etal-2024-language,kojima-etal-2024-multilingual}, and culture neurons~\cite{ying-etal-2025-disentangling,namazifard-poech-2025-isolating}.

The integrated-gradient formulation requires multiple forward/backward passes per example, which is expensive when one wants to score every neuron in a 4B--20B-parameter LLM.
Following \cite{yang-etal-2024-mitigating}, we therefore adopt the first-order Taylor approximation of the gradient around a zero reference activation, which reduces the score to a single ``gradient $\times$ activation'' product computable with one backward pass.
Concretely, given an input $x$, and a target output $y$, the attribution score of the $i$-th neuron with intermediate activation $h_i$ is defined as
\begin{equation}
A_i^{(x, y)} \;=\; h_i \cdot \frac{\partial P(y \mid x)}{\partial h_i}.
\label{eq:attribution}
\end{equation}
Equation~\eqref{eq:attribution} can be interpreted as a first-order approximation of the drop in the target probability that would result from ablating neuron $i$ to zero, and is therefore a tractable proxy for the causal importance of the neuron.

\subsection{Identifying task-specific influential neurons and suppressing them}
\label{subsec:method_amplify}

\paragraph{Score aggregation on a training split.}
Let $\mathcal{D}_\text{train} = \{(x^{(j)}, y^{(j)})\}_{j=1}^{N}$ be the training split of a target domain (e.g., a legal benchmark).
For each neuron $i$ we aggregate Eq.~\eqref{eq:attribution} over the training examples to obtain a domain-level importance score
\begin{equation}
\bar{A}_i \;=\; \frac{1}{N} \sum_{j=1}^{N} A_i^{(x^{(j)}, y^{(j)})},
\label{eq:agg_attribution}
\end{equation}
and select the top-$k$ neurons $\mathcal{N}_k = \arg\text{top-}k_i \, \bar{A}_i$ with the highest aggregated scores.
Because $\mathcal{N}_k$ is determined entirely from $\mathcal{D}_\text{train}$, the test split remains untouched and is used only for evaluation.

\paragraph{Activation suppression at inference.}
Standard prior work uses attribution scores either to \emph{suppress} (set to zero) or to qualitatively \emph{amplify} the selected neurons in order to verify causal effects~\cite{dai-etal-2022-knowledge}.
In this study, we suppress the neuronal activity by suppressing the activations of the selected neurons by $\gamma = 0$,
\begin{equation}
\tilde{h}_i \;=\;
\begin{cases}
\gamma \cdot h_i & \text{if } i \in \mathcal{N}_k, \\
h_i & \text{otherwise},
\end{cases}
\label{eq:amplify}
\end{equation}
while leaving all other neurons and all model parameters unchanged.

\paragraph{Random-neuron control.}
A natural concern is that any modest perturbation of the network might happen to improve scores on a benchmark.
To rule this out, we always compare against a control in which the same number $k$ of neurons is selected uniformly at random and suppressed by the same factor $\gamma$.
A genuine effect of attribution should therefore manifest as a systematic gap between the top-$k$ amplification and the random-amplification baseline.

\subsection{Application to legal reasoning and other tasks}
\label{subsec:neuron_attribution}

To investigate legal specific neurons in transformer-based language models, we use three legal reasoning datasets of three different legal jurisdictions of China, the US, and Switzerland by using LexEval~\cite{lexeval2024}, LexGLUE~\cite{chalkidis-etal-2022-lexglue}, and LEXam~\cite{fan2025lexam}.
LexEval covers a wide variety of Chinese legal questions, LexGLUE and LEXam offer Swiss legal questions in English and German.
In long-form legal question answering of legal reasoning, multiple valid answer formulations may exist to express the same logical reasoning; no definitive evaluation methodology has yet been established, and the best evaluation method of the open-ended legal question answering task is yet under discussion.
LLM-as-a-Judge is not sufficient because it sometimes performs worse than human evaluation in the legal domain.
Accordingly, we target only multiple-choice style reasoning tasks among LexEval, CaseHOLD task of LexGLUE, and  LEXam datasets.
In order to compare whether the observed phenomenon is unique to the legal domain or common across other application domain reasoning tasks, we use the mathematical reasoning dataset GSM8K~\cite{cobbe2021gsm8k}, the medical domain specific task MedQA~\cite{jin2021disease}, and the spatial reasoning task PIQA~\cite{Bisk2020}. 
To compare the difference of reasoning tasks and general translation tasks, we use WMT19~\cite{barrault-etal-2019-findings}.

\begin{table*}[t!]
    \centering
    \small\begin{tabular}{lc|ccccccc}
            \toprule
             \multicolumn{2}{c|}{\bf Model}  & \bf LexEval & \bf LexGLUE & \bf LEXam & \bf GSM8K & \bf MedQA & \bf PIQA & \bf WMT19\\
            \midrule
            \multicolumn{2}{c|}{Language} & ZH & EN & EN, DE & EN & EN, ZH & EN & multi  \\
            \midrule
            \multirow{3}{*}{\makecell{Qwen3-8B}}
& base & 64.52 & 67.80 & 14.20 &  80.14 & 68.21 & 78.78 & 88.44 \\
& identified & 0.0 & 0.20 & 1.40 & 0.76 & 0.29 & 0.0 & 0.0 \\
& random & 59.94 & 65.00 & 43.60 & 74.91 & 69.39 & 72.58 & 86.67 \\
\midrule
            \multirow{3}{*}{\makecell{Qwen3-14B}}
& base & 68.37 & 70.80 & 24.00 & 86.35 & 79.35 & 87.49 & 90.22  \\
& identified & 0.10 & 0.20 & 4.60 & 0.23 & 0.47 & 0.76 & 0.00 \\
& random & 66.49 & 71.00 & 31.60 & 84.69 & 76.90 & 84.11 & 89.33 \\
\midrule
            \multirow{3}{*}{\makecell{Gemma-3-12b-it}}
& base & 28.51 & 47.29 & 39.80  & 85.90 & 61.32 & 82.37 & 80.67 \\
& identified & 0.42 & 7.20 & 1.20 & 0.08 & 0.16 & 1.41 & 0.0 \\
& random & 51.61 & 62.60 & 7.40 & 77.18 & 58.07 & 82.15 & 75.00 \\
\midrule
            \multirow{3}{*}{\makecell{Gemma-4-E4B-it}}
& base & 2.39 & 11.80 & 18.20 & 79.08 & 30.30 & 0.76 & 43.78 \\
& identified & 0.10 & 1.40 & 1.40 & 1.06 & 0.00 & 0.00 & 0.11 \\
& random  & 6.45 & 37.40 & 14.80 & 74.98 & 19.50 & 0.22 & 25.00 \\
\midrule
            \multirow{3}{*}{\makecell{GPT-OSS-20B}}
& base & 6.24 & 20.00 & 20.80  & 62.09 & 5.04 & 11.21 & 43.33 \\
& identified & 0.0 & 15.0 & 0.0 & 0.15 & 0.0 & 23.07 & 12.67 \\
& random & 18.31 & 46.60 & 13.0 & 75.28 & 23.04 & 38.85 & 41.67 \\
 \midrule
            \multirow{3}{*}{\makecell{Olmo-3-1025-7B}}
& base & 13.42 & 48.80 & 18.20 & 52.77 & 35.81 & 73.34 & 38.44 \\
& identified & 0.52 & 4.20 & 0.0 & 1.21 & 0.82 & 2.29 & 0.33 \\
& random  & 37.36 & 60.80 & 18.20 & 32.98 & 34.42 & 74.43 & 38.67 \\
\midrule
            \multirow{3}{*}{\makecell{Llama-3.1-8B-Instruct}}
& base & 24.97 & 53.40 & 35.00  & 37.00 & 66.90 & 83.03 & 74.22 \\
& identified & 0.0 & 0.0 & 0.0 & 0.0 & 0.02 & 0.0 & 0.0 \\
& random  & 46.41 & 64.40 & 9.80 & 25.02 & 50.41 & 79.98 & 71.89 \\

        \bottomrule
    \end{tabular}
    \caption{Results of task-wise identified neuron suppression. ``base'' is the original model results while ``identified'' and ``random'' are neurons suppressed results. Identifies neurons are at top 0.5\% of all in each model.}
    \label{tab:general_result_top0.5per}
\end{table*}

\paragraph{Experimental settings}
We perform our experiments on seven open-weight transformer-based reasoning models with different pretraining data and size: Qwen3-8B, Qwen3-14B~\cite{qwen3technicalreport}, Gemma-3-12b-it~\cite{gemma_2025}, Gemma-4-E4B-it, GPT-OSS-20B~\cite{openai2025gptoss120bgptoss20bmodel}, Olmo-3-1025-7B~\cite{olmo2026olmo3}, Llama-3.1-8B-Instruct.
We split 7 datasets into training and test sets and identified important neurons using the training split, then measured task-specific accuracy on the test split.
GSM8K was evaluated using LightEval~\cite{lighteval}, based on eight generations produced with chain-of-thought prompting while WMT19 was evaluated by the F1 score of generated texts to the gold answers.
For the multiple-choice tasks other than GSM8K and WMT19, accuracy was evaluated against the gold answers.

\paragraph{Criteria of influential neurons from neuron attribution scores}
Whereas prior neuron extraction methods typically identified neurons whose scores exceeded a fixed threshold, we note that such thresholds may depend on the specific task. 
Because our goal is to analyze influential neurons across tasks, a threshold-based approach may obscure neurons that play an important role in particular tasks, especially when the corresponding datasets are limited in size.
We therefore adopt a top-n\% selection strategy for each task after computing neuron attribution scores. Specifically, we extract neurons at three selection ratios, top 1\%, 0.5\%, and 0.3\%, and find that task-specific neurons are most clearly observed around the top 0.5\% level. The results of suppressing neurons selected at the top 1\% and top 0.3\% levels are provided in the Appendix \ref{sec:top1and0.3_neuron_suppression_appendix}.
\section{Result}

\begin{figure*}[t]
    \centering
    \includegraphics[width=15cm]{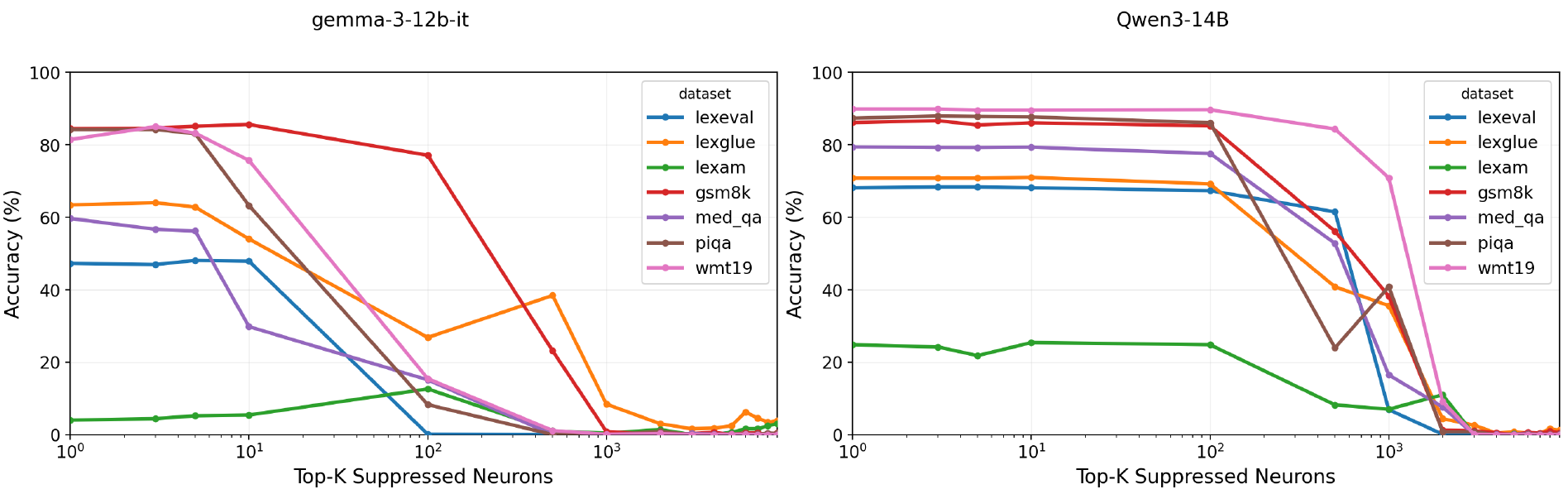}
    \caption{The top K neuron suppression result of Gemma-3-12b-it and Qwen-3-14B sorted by neuron attribution score.}
    \label{fig:topk_suppress}
\end{figure*}

\subsection{Task-wise influential neurons}

\paragraph{Identified neurons surely contribute to task performance}
Following prior studies based on neuron attribution scores, we verified that the neurons identified within each task by neuron attribution scores are genuinely influential for task performance.
Table~\ref{tab:general_result_top0.5per} reports the results for the baseline, suppression of the selected neurons, and suppression of randomly selected neurons.
The baseline results are the averages over seven inference runs.
By suppressing the neuron outputs to 0 in identified neurons, model accuracy drops to nearly zero. 
By contrast, even when the same number of neurons are randomly suppressed, task performance remains almost identical to the baseline.
This experiment suggests that the top 0.5\% of extracted neurons are highly influential to the task performance. 

\paragraph{Score change by top k neuron suppression}
We analyze the relationship between the number of suppressed neurons and task performance for the neurons identified in each task.
Figure~\ref{fig:topk_suppress} shows the changes in task accuracy when the top-k neurons are suppressed.
We observe a clear decline in accuracy when suppressing roughly the top 100 to top 1,000 neurons. 
By contrast, accuracy does not decrease substantially when only the top 1 to top 100 important neurons are suppressed.
The remaining results for top-k neuron suppression are provided in Appendix~\ref{sec:score_change_by_topk_neuron_suppression_appendix}.

\subsection{Influential neurons across tasks}
\begin{figure*}[t]
    \centering
    \includegraphics[width=16cm]{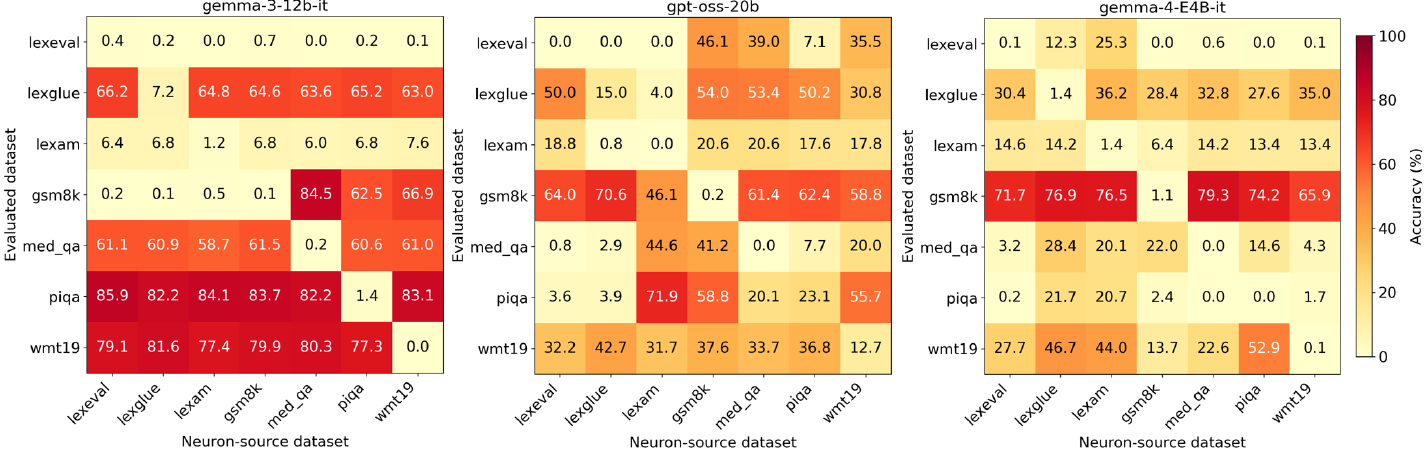}
    \caption{Task-wise neuron suppression results in  Gemma-3-12B-it, Gemma-4-E4B-it, GPT-OSS-20B.}
    \label{fig:neuron_suppress_0.5}
\end{figure*}

Next, we examine whether the neurons identified for each task are specifically influential for that task or remain influential across other tasks as well. To this end, we conduct an experiment in which we suppress the top 0.5\% of influential neurons identified for one task and then evaluate performance on the remaining tasks. If neurons that are important for a given task affect only that task, suppressing them should not alter accuracy when the model is evaluated on other tasks.
Figure~\ref{fig:neuron_suppress_0.5} presents the results of this cross-task suppression analysis across the seven tasks.
In Gemma-3-12B-it, GPT-OSS-20B, and Gemma-4-E4B-it, suppressing the influential neurons computed from one task tend not to affect for other task performances.
In contrast, when the same top 0.5\% neurons of one task are suppressed, and the task accuracy is evaluated for that task, the model performance drops nearly 0. 
These results suggest that, for these three models, our method successfully extracts task-specific neurons.
In Olmo-3-1025-7B and Llama-3.1-8B-Instruct, suppressing the influential neurons computed from GSM8K does not reduce model performance of other tasks, whereas performance drops sharply when the model's GSM8K accuracy is evaluated with the identified neurons of GSM8K. 
In Olmo-3-1025-7B, a similar trend is also observed for LexGLUE. 
However, for the remaining tasks, neuron suppression causes decline of nealy all task accuracy to approach zero.
For Qwen3-8B and Qwen3-14B, suppressing the identified neurons causes performance on nearly all tasks to drop close to zero, making it difficult to conclude that domain-specific neurons have been successfully isolated. Comprehensive experimental results are provided in Appendix~\ref{sec:top0.5_scored_neuron_suppression_appendix}.

\begin{figure*}[t]
    \centering
    \includegraphics[width=16cm]{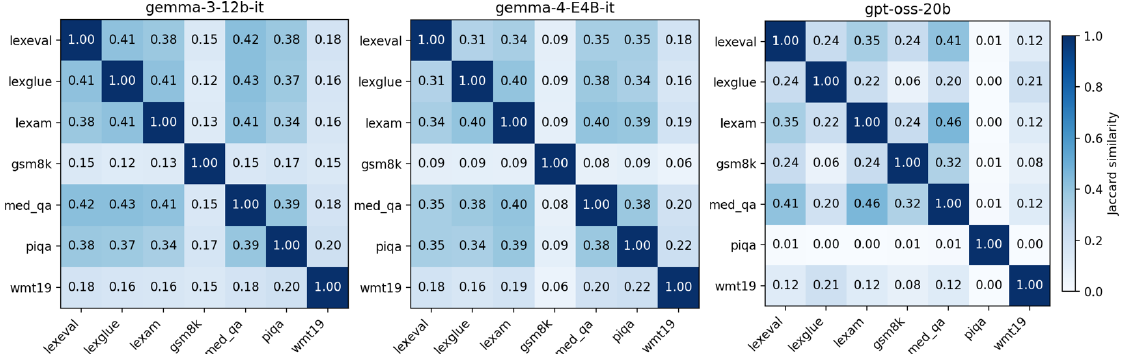}
    \caption{Influential neuron overlaps across tasks in  Gemma-3-12B-it, Gemma-4-E4B-it, and GPT-OSS-20B.}
    \label{fig:neuron_overlap}
\end{figure*}
\begin{table}[t]
\centering
  \footnotesize
\begin{tabular}{lrr}
\toprule
\textbf{Model} & \textbf{\#Top0.5\%} & \textbf{\#Shared} \\
\midrule
Qwen3-8B & 5,529 &  316 \\
Qwen3-14B & 8,396 & 474 \\
 Gemma-3-12B-it & 9,338 &  1,310 \\
Gemma-4-E4B-it & 4,945 & 359 \\
GPT-OSS-20B & 1,305 & 0  \\
Olmo-3-1025-7B  & 5,488 & 1,078  \\
Llama-3.1-8B-Instruct & 5,570 &  2,055 \\
\bottomrule
\end{tabular}
\caption{Number of neurons at top 0.5\% and the number of the shared neurons across seven tasks.}
\label{table:shared_neuron_top0.5per}
\end{table}

\begin{figure*}[t]
    \centering
    \includegraphics[width=15.5cm]{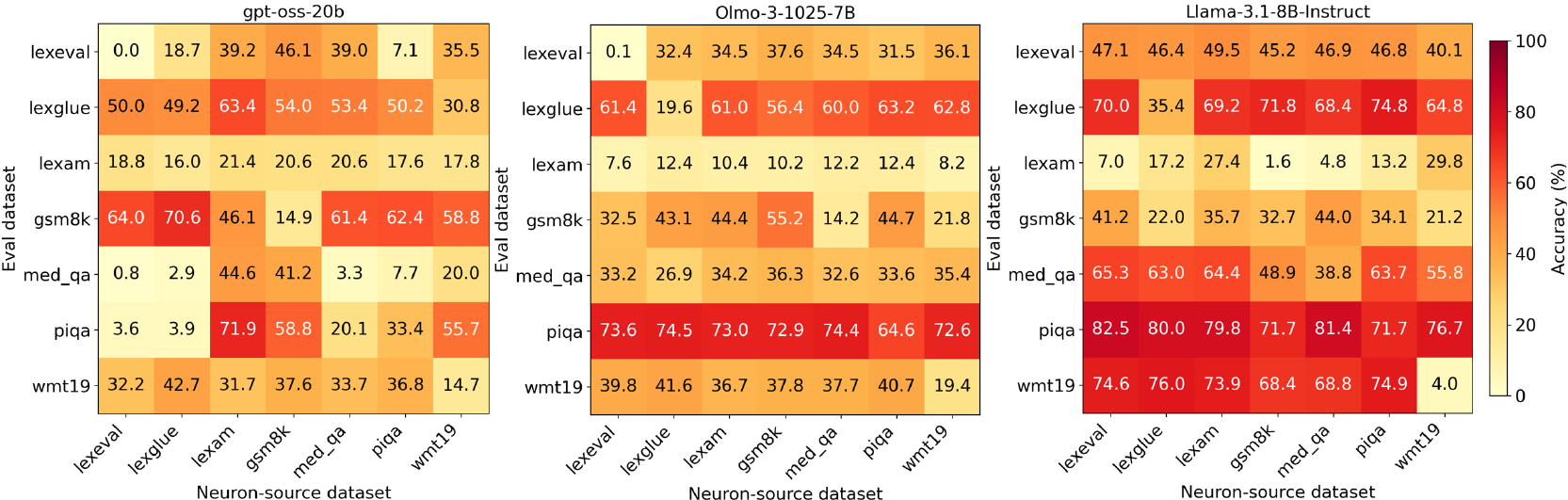}
    \caption{Neuron suppression results when cross-task influential neurons are removed in GPT-OSS-20B, Olmo-3-1025-7B, and Llama-3.1-8B-Instruct.}
    \label{fig:top_only_neurons_0.5per}
\end{figure*}

\paragraph{Do task-wise influential neurons overlap across tasks?}
Thus far, we have examined the properties of influential neurons identified for a specific task. 
We now turn to neurons that are influential across multiple tasks. 
In particular, we investigate whether neurons identified as influential for one task are also influential for others, or whether there exist neurons specific to a particular task or domain, such as legal or mathematical reasoning.
Table~\ref{table:shared_neuron_top0.5per} shows the number of top 0.5\% neurons and the number of neurons that are identified as influential across all seven tasks. We refer to these cross-task influential neurons as shared neurons.
When 0.5\% of neurons are selected at random, the probability that a given neuron is selected for all seven tasks should be $0.005^7$, which is vanishingly small. From this perspective, the number of shared neurons reported in Table~\ref{table:shared_neuron_top0.5per} appears to be relatively large.
Figure~\ref{fig:neuron_overlap} visualizes the overlap of influential neurons among the seven tasks. 
In Figure~\ref{fig:neuron_overlap}, at least 5\% of the selected neurons are shared across tasks in  Gemma-3-12B-it and gemma-4-E4B-it, whereas PIQA in GPT-OSS-20B exhibits almost no overlap with the selected neurons of other tasks. 
The complete results of neuron overlap are provided in Appendix~\ref{sec:top0.5_scored_neuron_suppression_appendix}.

\paragraph{Truly task-specific neurons via removal of cross-task influential neurons.}
\label{sec:important_neuron_shared_neuron_removal}
To isolate neurons that are highly important for individual tasks, we attempt to extract task-specific neurons by removing the shared neurons that are identified as influential across all seven tasks, from each task's list of the top 0.5\% identified neurons.
Figure~\ref{fig:top_only_neurons_0.5per} shows the neuron suppression results of each dataset.
In Gemma-3-12B-it, Gemma-4-E4B-it, and GPT-OSS-20B, suppressing the top 0.5\% of neurons ranked by attribution score reduces accuracy only on the task from which those neurons were identified, while leaving the other tasks largely unaffected. A similar tendency emerges in Olmo-3-1025-7B, Llama-3.1-8B-Instruct, Qwen3-8B, and Qwen3-14B once the neurons shared across all seven datasets are removed: accuracy declines only on the task from which the neurons were identified, while remaining largely unchanged on the others. These results suggest that some models contain neurons that play a general role across all tasks, and that task-specific neurons can be recovered by removing such broadly important neurons from the top-ranked set. The remaining results of the identified neuron distributions are provided in Appendix~\ref{sec:important_neuron_shared_neuron_removal_appendix}.

\subsection{Domain-specific influential neurons}
\paragraph{Characteristics of task-specific neurons in the legal domain.}
Finally, we return to the task-specific analysis of influential neurons and examine the characteristics of legal-domain influential neurons. 
As shown by the cross-suppression results in Figure~\ref{fig:neuron_suppress_0.5},  the identified neurons also behave as task-specific neurons in GPT-OSS-20B, Gemma-3-12B-it, and gemma-4-E4B-it. 
In these three models, we observe a weak tendency across the three legal datasets (LexEval, LexGLUE, and LEXam), namely, suppressing the influential neurons for one legal task degrades performance on the other two as well. 
Consistent with this, the overlap of identified neurons shown in Figure~\ref{fig:neuron_overlap} indicates that the degree of overlap among task-associated neurons is relatively high among LexEval, LexGLUE, and LEXam, as seen in the upper-left portion of the figure. 
This phenomenon is observed in Llama-3.1-8B-Instruct, Olmo-3-1025-7B, Gemma-3-12B-it, gemma-4-E4B-it, and GPT-OSS-20B.

\paragraph{The relationship between legal task-specific neurons and other task-specific neurons.}
The relationship between legal task-specific neurons and other task-specific neurons. Neurons associated with the legal datasets (LexEval, LexGLUE, and LEXam) exhibit substantial overlap with those identified for MedQA and PIQA in Llama-3.1-8B-Instruct, Gemma-3-12B-it, gemma-4-E4B-it, and Olmo-3-1025-7B. In GPT-OSS-20B, they instead overlap substantially with the neurons identified for GSM8K and MedQA. These findings suggest the existence of neurons that are broadly important across reasoning tasks.

\begin{figure*}[t]
    \centering
    \includegraphics[width=15cm]{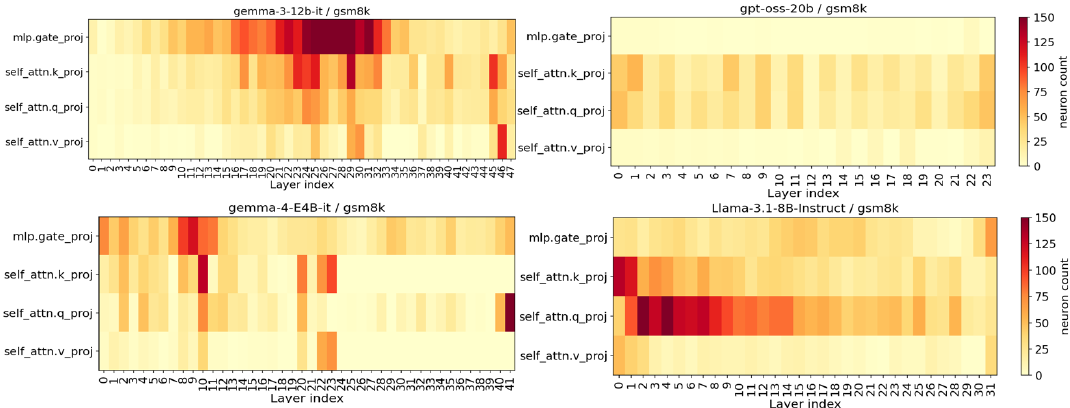}
    \caption{Important neuron distribution heatmap of gemma-3-12b, Gemma-4-E4B-it, GPT-OSS-20B, Llama-3.1-8B-Instruct in top 0.5\% neuron attribution score.}
    \label{fig:neuron_dist_heatmap}
    \vspace{-0.5em}
\end{figure*}

\section{Discussion}
\label{sec:discussion}
\paragraph{On which layers, MLPs, and attention modules are influential neurons distributed?}
A recurring line of inquiry has examined where the neurons most critical for LLM output generation reside within the network, from the first to the final layer, and whether neurons in the attention mechanism, where $\mathrm{Attention}(Q,K,V)=\mathrm{softmax}(QK^\top/\sqrt{d})V$ with $d$ denoting the key dimension, are as important as those in the MLP layers.
\citet{geva-etal-2021-transformer} showed that feed-forward layers function as key-value memories, in which the keys detect input patterns and the values encode next-token distributions, and that the model's final prediction is constructed in a bottom-up manner by composing these memories within each layer and iteratively refining them across layers.
Using causal mediation analysis, corrupting the subject's token embeddings and then restoring individual hidden states to their clean values, \citet{meng2022locating} found that mid-layer MLP modules at the subject's last token have the strongest indirect causal effect on factual predictions, with this effect disappearing when MLP computations are severed but not when attention is severed, implicating mid-layer MLPs as the locus of factual recall.
\citet{yamamoto2026neuronlevel} similarly found that most identified cultural neurons reside in the shallow to middle MLP modules, and that this trend is consistent across models.
In contrast, \citet{nikankin2026reasoningmodelsknowwhats} trained separate probes on the residual-stream activations at specific token positions for each generation step in order to localize where importance is encoded. They found that probes trained on individual layers achieved comparable accuracy, suggesting that the components critical to the model's output are not localized to any particular layer but are instead broadly distributed across the network.

In our experiment, the identified task specific neurons distribution of Gemma-3-12b-it and Gemma-4-E4B-it are similar to existing analyses that important neurons are placed in the middle of the MLP layer, while Llama-3.1-8B-Instruct and GPT-OSS-20B seem to have important neurons in key $K$ and query $Q$ of attention across from the shallow to the deep layer is shown in Figure~\ref{fig:neuron_dist_heatmap}.
Appendix~\ref{sec:important_neuron_distribution_appendix} for the full distribution of top influential neurons across all tasks and models.
Appendix~\ref{sec:important_neuron_shared_neuron_removal_appendix} shows the distribution of neurons selected via removal of cross-task influential neurons from identified neurons, and this also shows a similar tendency.

\paragraph{Middle-MLP localization may not generalize to long, complex inputs.}
The distribution of important neurons varies across models, while within each model, a broadly consistent pattern is observed across all analyzed tasks. In Gemma-3-12B-it and Gemma-4-E4B-it, all seven tasks show a mild tendency for important neurons to be distributed around the middle MLP layers, consistent with prior hypotheses. In contrast, Olmo-3-1025-7B and Llama-3.1-8B-Instruct exhibit a concentration in the attention $K$ and $Q$ components from layer 0 through roughly the first two-thirds of the network. Meanwhile, in Qwen3-8B and Qwen3-14B, important neurons are distributed broadly across all layers in the MLP, $K$, and $Q$ components, but not in projection component $V$.
The rest result of identified neuron distributions are shown in Appendix~\ref{sec:important_neuron_distribution_appendix}.
In prior studies that identified important neurons in the middle layer of MLPs~\cite{meng2022locating, yamamoto2026neuronlevel}, they used short, single-text inputs for relatively general knowledge tasks, such as multilingualism, bias, and culture understanding. 
However, when examining neuron activation scores in more applied domains such as mathematics and coding studied by \citet{nikankin2026reasoningmodelsknowwhats}, prior findings may not be applicable. 
Reasoning tasks in domains such as mathematics, law, and medicine involve complex logical structures spanning multiple sentences, unlike those considered in prior work. 
The hypothesis that the middle layers of the MLP are critical for text generation was task-dependent rather than a general principle of LLMs, because all domain applications, including mathematics, medicine, and law, involve relatively long input text and complex text structure.
Therefore, our setting differs from previous studies that analyzed single sentences and found that the MLP layers at the final token of the subject play an important role~\cite{meng2022locating, yamamoto2026neuronlevel}. 
This difference may explain why the distribution of influential neurons observed in our study does not align with prior findings.

\section{Conclusion}

We presented a neuron-level analysis of legal-domain reasoning in LLMs, comparing it with other applied domains across seven open-weight models.
Using neuron attribution scores to rank and suppress influential neurons, we confirmed that suppressing the identified neurons collapses accuracy on the target task, whereas suppressing the same number of random neurons does not.
We found a small subset of neurons influential across seven tasks; once these are removed, suppressing the remaining neurons degrades only the task from which they were identified, revealing genuinely task-specific neurons in every model studied.
Within the legal domain, the three benchmarks exhibit relatively high neuron overlap and tend to be affected jointly, suggesting legal components that span jurisdictions.
Finally, influential neurons do not consistently localize to middle MLP layers, indicating that this widely held hypothesis, established mainly on short and general-knowledge inputs, may be task- and input-format-dependent rather than universal.
The findings of this study suggest the existence of neurons responsible for domain-specific reasoning. 
This indicates that developing training methods that selectively target such neurons may contribute to improving the performance of LLMs in specific domains.

\section*{Acknowledgments}
This work was supported by the “R\&D Hub Aimed at Ensuring Transparency and Reliability of Generative AI Models” project of the Ministry of Education, Culture, Sports, Science and Technology, and JST ACT-X JPMJAX25C6 \& JPMJAX25CN.

\section*{Limitations}
In this paper, to identify legal-domain-specific neurons, we analyze all existing datasets that require legal reasoning ability, using LexEval, LexGLUE, and LEXam to cover China, U.S., and Switzerland law. 
However, because laws are, in principle, effective only within each country’s own jurisdiction, and because the language in which they are expressed also differs across countries, it would be desirable for additional datasets to be proposed and evaluated that measure legal reasoning ability across a broader range of jurisdictions and languages.

\section*{Ethics Statement}
This work focuses on analyzing internal neuron activations in large reasoning models to better understand and improve inference efficiency. Our analysis is conducted on publicly available benchmark datasets spanning legal reasoning and understanding (LexEval, LexGLUE, LEXam), mathematical reasoning (GSM8K), medical question answering (MedQA), physical commonsense reasoning (PIQA), and machine translation (WMT19). 
The study does not involve human subjects, the collection of personal data, or user-generated content, and all datasets are used in accordance with their respective licenses and intended research purposes.
We do not anticipate direct negative societal impacts from the proposed method itself.
While our method aims to improve model efficiency through neuron activation analysis, we acknowledge that interpretability techniques applied to domain-specific benchmarks, particularly in sensitive areas such as medical and legal reasoning, should not be interpreted as endorsements of deploying such models in high-stakes decision-making without appropriate expert oversight.


\bibliography{custom}

@inproceedings{query_relavant_neuron,
author = {Chen, Lihu and Dejl, Adam and Toni, Francesca},
title = {Identifying query-relevant neurons in large language models for long-form texts},
year = {2025},
isbn = {978-1-57735-897-8},
publisher = {AAAI Press},
url = {https://doi.org/10.1609/aaai.v39i22.34529},
doi = {10.1609/aaai.v39i22.34529},
booktitle = {Proceedings of the Thirty-Ninth AAAI Conference on Artificial Intelligence and Thirty-Seventh Conference on Innovative Applications of Artificial Intelligence and Fifteenth Symposium on Educational Advances in Artificial Intelligence},
articleno = {2631},
numpages = {10},
series = {AAAI'25/IAAI'25/EAAI'25}
}

@inproceedings{yang-etal-2024-mitigating,
    title = "Mitigating Biases for Instruction-following Language Models via Bias Neurons Elimination",
    author = "Yang, Nakyeong  and
      Kang, Taegwan  and
      Choi, Stanley Jungkyu  and
      Lee, Honglak  and
      Jung, Kyomin",
    editor = "Ku, Lun-Wei  and
      Martins, Andre  and
      Srikumar, Vivek",
    booktitle = "Proceedings of the 62nd Annual Meeting of the Association for Computational Linguistics (Volume 1: Long Papers)",
    month = aug,
    year = "2024",
    address = "Bangkok, Thailand",
    publisher = "Association for Computational Linguistics",
    url = "https://aclanthology.org/2024.acl-long.490/",
    doi = "10.18653/v1/2024.acl-long.490",
    pages = "9061--9073"
}

@inproceedings{kojima-etal-2024-multilingual,
    title = "On the Multilingual Ability of Decoder-based Pre-trained Language Models: Finding and Controlling Language-Specific Neurons",
    author = "Kojima, Takeshi  and
      Okimura, Itsuki  and
      Iwasawa, Yusuke  and
      Yanaka, Hitomi  and
      Matsuo, Yutaka",
    editor = "Duh, Kevin  and
      Gomez, Helena  and
      Bethard, Steven",
    booktitle = "Proceedings of the 2024 Conference of the North American Chapter of the Association for Computational Linguistics: Human Language Technologies (Volume 1: Long Papers)",
    month = jun,
    year = "2024",
    address = "Mexico City, Mexico",
    publisher = "Association for Computational Linguistics",
    url = "https://aclanthology.org/2024.naacl-long.384/",
    doi = "10.18653/v1/2024.naacl-long.384",
    pages = "6919--6971"
}

@inproceedings{ying-etal-2025-disentangling,
    title = "Disentangling Language and Culture for Evaluating Multilingual Large Language Models",
    author = "Ying, Jiahao  and
      Tang, Wei  and
      Zhao, Yiran  and
      Cao, Yixin  and
      Rong, Yu  and
      Zhang, Wenxuan",
    editor = "Che, Wanxiang  and
      Nabende, Joyce  and
      Shutova, Ekaterina  and
      Pilehvar, Mohammad Taher",
    booktitle = "Proceedings of the 63rd Annual Meeting of the Association for Computational Linguistics (Volume 1: Long Papers)",
    month = jul,
    year = "2025",
    address = "Vienna, Austria",
    publisher = "Association for Computational Linguistics",
    url = "https://aclanthology.org/2025.acl-long.1082/",
    doi = "10.18653/v1/2025.acl-long.1082",
    pages = "22230--22251",
    ISBN = "979-8-89176-251-0"
}

@inproceedings{namazifard-poech-2025-isolating,
    title = "Isolating Culture Neurons in Multilingual Large Language Models",
    author = "Namazifard, Danial  and
      Galke Poech, Lukas",
    editor = "Inui, Kentaro  and
      Sakti, Sakriani  and
      Wang, Haofen  and
      Wong, Derek F.  and
      Bhattacharyya, Pushpak  and
      Banerjee, Biplab  and
      Ekbal, Asif  and
      Chakraborty, Tanmoy  and
      Singh, Dhirendra Pratap",
    booktitle = "Proceedings of the 14th International Joint Conference on Natural Language Processing and the 4th Conference of the Asia-Pacific Chapter of the Association for Computational Linguistics",
    month = dec,
    year = "2025",
    address = "Mumbai, India",
    publisher = "The Asian Federation of Natural Language Processing and The Association for Computational Linguistics",
    url = "https://aclanthology.org/2025.findings-ijcnlp.45/",
    doi = "10.18653/v1/2025.findings-ijcnlp.45",
    pages = "768--785",
    ISBN = "979-8-89176-303-6"
}

@inproceedings{fan2025lexam,
title={{LEX}am: Benchmarking Legal Reasoning on 340 Law Exams},
author={Yu Fan and Jingwei Ni and Jakob Merane and Yang Tian and Yoan Hermstrüwer and Yinya Huang and Mubashara Akhtar and Etienne Salimbeni and Florian Geering and Oliver Dreyer and Daniel Brunner and Markus Leippold and Mrinmaya Sachan and Alexander Stremitzer and Christoph Engel and Elliott Ash and Joel Niklaus},
booktitle={The Fourteenth International Conference on Learning Representations},
year={2026},
url={https://openreview.net/forum?id=xNhbMyXsJn}
}

@misc{openai2025gptoss120bgptoss20bmodel,
      title={gpt-oss-120b \& gpt-oss-20b Model Card}, 
      author={OpenAI},
      year={2025},
      eprint={2508.10925},
      archivePrefix={arXiv},
      primaryClass={cs.CL},
      url={https://arxiv.org/abs/2508.10925}, 
}

@article{gemma_2025,
    title={Gemma 3 Technical Report},
    url={https://goo.gle/Gemma3Report},
    author={{Gemma Team}},
    year={2025}
}

@misc{qwen3technicalreport,
      title={Qwen3 Technical Report}, 
      author={{Qwen Team}},
      year={2025},
      eprint={2505.09388},
      archivePrefix={arXiv},
      primaryClass={cs.CL},
      url={https://arxiv.org/abs/2505.09388}, 
}

@misc{cobbe2021gsm8k,
      title={Training Verifiers to Solve Math Word Problems}, 
      author={Karl Cobbe and Vineet Kosaraju and Mohammad Bavarian and Mark Chen and Heewoo Jun and Lukasz Kaiser and Matthias Plappert and Jerry Tworek and Jacob Hilton and Reiichiro Nakano and Christopher Hesse and John Schulman},
      year={2021},
      eprint={2110.14168},
      archivePrefix={arXiv},
      primaryClass={cs.LG},
      url={https://arxiv.org/abs/2110.14168}, 
}

@article{jin2021disease,
author = {Jin, Di and Pan, Eileen and Oufattole, Nassim and Weng, Wei-Hung and Fang, Hanyi and Szolovits, Peter},
title = {What Disease Does This Patient Have? A Large-Scale Open Domain Question Answering Dataset from Medical Exams},
JOURNAL = {Applied Sciences},
VOLUME = {11},
YEAR = {2021},
NUMBER = {14},
ARTICLE-NUMBER = {6421},
url = {https://www.mdpi.com/2076-3417/11/14/6421},
ISSN = {2076-3417},
doi = {10.3390/app11146421}
}

@inproceedings{lexeval2024,
title={LexEval: A Comprehensive Chinese Legal Benchmark for Evaluating Large Language Models},
author={Haitao Li and You Chen and Qingyao Ai and Yueyue Wu and Ruizhe Zhang and Yiqun LIU},
booktitle={The Thirty-eight Conference on Neural Information Processing Systems Datasets and Benchmarks Track},
year={2024},
url={https://openreview.net/forum?id=8RaxRs5VDf}
}

@inproceedings{sundararajan2017axiomatic,
author = {Sundararajan, Mukund and Taly, Ankur and Yan, Qiqi},
title = {Axiomatic attribution for deep networks},
year = {2017},
publisher = {JMLR.org},
booktitle = {Proceedings of the 34th International Conference on Machine Learning - Volume 70},
pages = {3319–3328},
numpages = {10},
location = {Sydney, NSW, Australia},
series = {ICML'17}
}

@inproceedings{dhamdhere2018important,
title={How Important is a Neuron?},
author={Kedar Dhamdhere and Mukund Sundararajan and Qiqi Yan},
booktitle={International Conference on Learning Representations},
year={2019},
url={https://openreview.net/forum?id=SylKoo0cKm},
}

@misc{nikankin2026reasoningmodelsknowwhats,
      title={Reasoning Models Know What's Important, and Encode It in Their Activations}, 
      author={Yaniv Nikankin and Martin Tutek and Tomer Ashuach and Jonathan Rosenfeld and Yonatan Belinkov},
      year={2026},
      eprint={2604.18307},
      archivePrefix={arXiv},
      primaryClass={cs.CL},
      url={https://arxiv.org/abs/2604.18307}, 
}

@inproceedings{geva-etal-2021-transformer,
    title = "Transformer Feed-Forward Layers Are Key-Value Memories",
    author = "Geva, Mor  and
      Schuster, Roei  and
      Berant, Jonathan  and
      Levy, Omer",
    editor = "Moens, Marie-Francine  and
      Huang, Xuanjing  and
      Specia, Lucia  and
      Yih, Scott Wen-tau",
    booktitle = "Proceedings of the 2021 Conference on Empirical Methods in Natural Language Processing",
    month = nov,
    year = "2021",
    address = "Online and Punta Cana, Dominican Republic",
    publisher = "Association for Computational Linguistics",
    url = "https://aclanthology.org/2021.emnlp-main.446/",
    doi = "10.18653/v1/2021.emnlp-main.446",
    pages = "5484--5495"
}

@inproceedings{chalkidis-etal-2022-lexglue,
    title = "{L}ex{GLUE}: A Benchmark Dataset for Legal Language Understanding in {E}nglish",
    author = "Chalkidis, Ilias  and
      Jana, Abhik  and
      Hartung, Dirk  and
      Bommarito, Michael  and
      Androutsopoulos, Ion  and
      Katz, Daniel  and
      Aletras, Nikolaos",
    editor = "Muresan, Smaranda  and
      Nakov, Preslav  and
      Villavicencio, Aline",
    booktitle = "Proceedings of the 60th Annual Meeting of the Association for Computational Linguistics (Volume 1: Long Papers)",
    month = may,
    year = "2022",
    address = "Dublin, Ireland",
    publisher = "Association for Computational Linguistics",
    url = "https://aclanthology.org/2022.acl-long.297/",
    doi = "10.18653/v1/2022.acl-long.297",
    pages = "4310--4330"
}

@inproceedings{Bisk2020,
  author       = {Yonatan Bisk and
                  Rowan Zellers and
                  Ronan Le Bras and
                  Jianfeng Gao and
                  Yejin Choi},
  title        = {{PIQA:} Reasoning about Physical Commonsense in Natural Language},
  booktitle    = {The Thirty-Fourth {AAAI} Conference on Artificial Intelligence, {AAAI}
                  2020, The Thirty-Second Innovative Applications of Artificial Intelligence
                  Conference, {IAAI} 2020, The Tenth {AAAI} Symposium on Educational
                  Advances in Artificial Intelligence, {EAAI} 2020, New York, NY, USA,
                  February 7-12, 2020},
  pages        = {7432--7439},
  publisher    = {{AAAI} Press},
  year         = {2020},
  url          = {https://doi.org/10.1609/aaai.v34i05.6239},
  doi          = {10.1609/AAAI.V34I05.6239},
  bibsource    = {dblp computer science bibliography, https://dblp.org}
}

@inproceedings{barrault-etal-2019-findings,
    title = "Findings of the 2019 Conference on Machine Translation ({WMT}19)",
    author = {Barrault, Lo{\"i}c  and
      Bojar, Ond{\v{r}}ej  and
      Costa-juss{\`a}, Marta R.  and
      Federmann, Christian  and
      Fishel, Mark  and
      Graham, Yvette  and
      Haddow, Barry  and
      Huck, Matthias  and
      Koehn, Philipp  and
      Malmasi, Shervin  and
      Monz, Christof  and
      M{\"u}ller, Mathias  and
      Pal, Santanu  and
      Post, Matt  and
      Zampieri, Marcos},
    editor = "Bojar, Ond{\v{r}}ej  and
      Chatterjee, Rajen  and
      Federmann, Christian  and
      Fishel, Mark  and
      Graham, Yvette  and
      Haddow, Barry  and
      Huck, Matthias  and
      Yepes, Antonio Jimeno  and
      Koehn, Philipp  and
      Martins, Andr{\'e}  and
      Monz, Christof  and
      Negri, Matteo  and
      N{\'e}v{\'e}ol, Aur{\'e}lie  and
      Neves, Mariana  and
      Post, Matt  and
      Turchi, Marco  and
      Verspoor, Karin",
    booktitle = "Proceedings of the Fourth Conference on Machine Translation (Volume 2: Shared Task Papers, Day 1)",
    month = aug,
    year = "2019",
    address = "Florence, Italy",
    publisher = "Association for Computational Linguistics",
    url = "https://aclanthology.org/W19-5301/",
    doi = "10.18653/v1/W19-5301",
    pages = "1--61"
}

@misc{olmo2026olmo3,
      title={Olmo 3}, 
      author={{Olmo Team}},
      year={2026},
      eprint={2512.13961},
      archivePrefix={arXiv},
      primaryClass={cs.CL},
      url={https://arxiv.org/abs/2512.13961}, 
}

@misc{cao2025jbeqajapanesebarexam,
      title={JBE-QA: Japanese Bar Exam QA Dataset for Assessing Legal Domain Knowledge}, 
      author={Zhihan Cao and Fumihito Nishino and Hiroaki Yamada and Nguyen Ha Thanh and Yusuke Miyao and Ken Satoh},
      year={2025},
      eprint={2511.22869},
      archivePrefix={arXiv},
      primaryClass={cs.CL},
      url={https://arxiv.org/abs/2511.22869}, 
}

@inproceedings{
yamamoto2026neuronlevel,
title={Neuron-Level Analysis of Cultural Understanding in Large Language Models},
author={Taisei Yamamoto and Ryoma Kumon and Danushka Bollegala and Hitomi Yanaka},
booktitle={The Fourteenth International Conference on Learning Representations},
year={2026},
url={https://openreview.net/forum?id=HZMmM3Dmri}
}

@misc{lighteval,
  author = {Habib, Nathan and Fourrier, Clémentine and Kydlíček, Hynek and Wolf, Thomas and Tunstall, Lewis},
  title = {LightEval: A lightweight framework for LLM evaluation},
  year = {2023},
  version = {0.11.0},
  url = {https://github.com/huggingface/lighteval}
}

@inproceedings{
rafailov2023direct,
title={Direct Preference Optimization: Your Language Model is Secretly a Reward Model},
author={Rafael Rafailov and Archit Sharma and Eric Mitchell and Christopher D Manning and Stefano Ermon and Chelsea Finn},
booktitle={Thirty-seventh Conference on Neural Information Processing Systems},
year={2023},
url={https://openreview.net/forum?id=HPuSIXJaa9}
}

@misc{deepseek-math,
      title={DeepSeekMath: Pushing the Limits of Mathematical Reasoning in Open Language Models}, 
      author={Zhihong Shao and Peiyi Wang and Qihao Zhu and Runxin Xu and Junxiao Song and Xiao Bi and Haowei Zhang and Mingchuan Zhang and Y. K. Li and Y. Wu and Daya Guo},
      year={2024},
      eprint={2402.03300},
      archivePrefix={arXiv},
      primaryClass={cs.CL},
      url={https://arxiv.org/abs/2402.03300}, 
}

@inproceedings{lightman2023lets,
title={Let's Verify Step by Step},
author={Hunter Lightman and Vineet Kosaraju and Yuri Burda and Harrison Edwards and Bowen Baker and Teddy Lee and Jan Leike and John Schulman and Ilya Sutskever and Karl Cobbe},
booktitle={The Twelfth International Conference on Learning Representations},
year={2024},
url={https://openreview.net/forum?id=v8L0pN6EOi}
}

@inproceedings{sun-etal-2025-reasonmed,
    title = "{R}eason{M}ed: A 370{K} Multi-Agent Generated Dataset for Advancing Medical Reasoning",
    author = "Sun, Yu  and
      Qian, Xingyu  and
      Xu, Weiwen  and
      Zhang, Hao  and
      Xiao, Chenghao  and
      Li, Long  and
      Zhao, Deli  and
      Huang, Wenbing  and
      Xu, Tingyang  and
      Bai, Qifeng  and
      Rong, Yu",
    editor = "Christodoulopoulos, Christos  and
      Chakraborty, Tanmoy  and
      Rose, Carolyn  and
      Peng, Violet",
    booktitle = "Proceedings of the 2025 Conference on Empirical Methods in Natural Language Processing",
    month = nov,
    year = "2025",
    address = "Suzhou, China",
    publisher = "Association for Computational Linguistics",
    url = "https://aclanthology.org/2025.emnlp-main.1344/",
    doi = "10.18653/v1/2025.emnlp-main.1344",
    pages = "26446--26467",
    ISBN = "979-8-89176-332-6"
}

@inproceedings{zhang-etal-2025-generation,
    title = "From Generation to Detection: A Multimodal Multi-Task Dataset for Benchmarking Health Misinformation",
    author = "Zhang, Zhihao  and
      Zhang, Yiran  and
      Zhou, Xiyue  and
      Huang, Liting  and
      Razzak, Imran  and
      Nakov, Preslav  and
      Naseem, Usman",
    editor = "Christodoulopoulos, Christos  and
      Chakraborty, Tanmoy  and
      Rose, Carolyn  and
      Peng, Violet",
    booktitle = "Findings of the Association for Computational Linguistics: EMNLP 2025",
    month = nov,
    year = "2025",
    address = "Suzhou, China",
    publisher = "Association for Computational Linguistics",
    url = "https://aclanthology.org/2025.findings-emnlp.1316/",
    doi = "10.18653/v1/2025.findings-emnlp.1316",
    pages = "24245--24260",
    ISBN = "979-8-89176-335-7"
}

@inproceedings{sviridova-etal-2024-casimedicos,
    title = "{C}asi{M}edicos-Arg: A Medical Question Answering Dataset Annotated with Explanatory Argumentative Structures",
    author = "Sviridova, Ekaterina  and
      Yeginbergen, Anar  and
      Estarrona, Ainara  and
      Cabrio, Elena  and
      Villata, Serena  and
      Agerri, Rodrigo",
    editor = "Al-Onaizan, Yaser  and
      Bansal, Mohit  and
      Chen, Yun-Nung",
    booktitle = "Proceedings of the 2024 Conference on Empirical Methods in Natural Language Processing",
    month = nov,
    year = "2024",
    address = "Miami, Florida, USA",
    publisher = "Association for Computational Linguistics",
    url = "https://aclanthology.org/2024.emnlp-main.1026/",
    doi = "10.18653/v1/2024.emnlp-main.1026",
    pages = "18463--18475"
}

@inproceedings{reddy-etal-2024-docfinqa,
    title = "{D}oc{F}in{QA}: A Long-Context Financial Reasoning Dataset",
    author = "Reddy, Varshini  and
      Koncel-Kedziorski, Rik  and
      Lai, Viet Dac  and
      Krumdick, Michael  and
      Lovering, Charles  and
      Tanner, Chris",
    editor = "Ku, Lun-Wei  and
      Martins, Andre  and
      Srikumar, Vivek",
    booktitle = "Proceedings of the 62nd Annual Meeting of the Association for Computational Linguistics (Volume 2: Short Papers)",
    month = aug,
    year = "2024",
    address = "Bangkok, Thailand",
    publisher = "Association for Computational Linguistics",
    url = "https://aclanthology.org/2024.acl-short.42/",
    doi = "10.18653/v1/2024.acl-short.42",
    pages = "445--458"
}

@inproceedings{magomere-etal-2025-finnli,
    title = "{F}in{NLI}: Novel Dataset for Multi-Genre Financial Natural Language Inference Benchmarking",
    author = "Magomere, Jabez  and
      Kochkina, Elena  and
      Mensah, Samuel  and
      Kaur, Simerjot  and
      Smiley, Charese",
    editor = "Chiruzzo, Luis  and
      Ritter, Alan  and
      Wang, Lu",
    booktitle = "Findings of the Association for Computational Linguistics: NAACL 2025",
    month = apr,
    year = "2025",
    address = "Albuquerque, New Mexico",
    publisher = "Association for Computational Linguistics",
    url = "https://aclanthology.org/2025.findings-naacl.257/",
    doi = "10.18653/v1/2025.findings-naacl.257",
    pages = "4545--4568",
    ISBN = "979-8-89176-195-7"
}

@inproceedings{
nikankin2025arithmetic,
title={Arithmetic Without Algorithms: Language Models Solve Math with a Bag of Heuristics},
author={Yaniv Nikankin and Anja Reusch and Aaron Mueller and Yonatan Belinkov},
booktitle={The Thirteenth International Conference on Learning Representations},
year={2025},
url={https://openreview.net/forum?id=O9YTt26r2P}
}

@inproceedings{tang-etal-2024-language,
    title = "Language-Specific Neurons: The Key to Multilingual Capabilities in Large Language Models",
    author = "Tang, Tianyi  and
      Luo, Wenyang  and
      Huang, Haoyang  and
      Zhang, Dongdong  and
      Wang, Xiaolei  and
      Zhao, Xin  and
      Wei, Furu  and
      Wen, Ji-Rong",
    editor = "Ku, Lun-Wei  and
      Martins, Andre  and
      Srikumar, Vivek",
    booktitle = "Proceedings of the 62nd Annual Meeting of the Association for Computational Linguistics (Volume 1: Long Papers)",
    month = aug,
    year = "2024",
    address = "Bangkok, Thailand",
    publisher = "Association for Computational Linguistics",
    url = "https://aclanthology.org/2024.acl-long.309/",
    doi = "10.18653/v1/2024.acl-long.309",
    pages = "5701--5715"
}

@inproceedings{dai-etal-2022-knowledge,
    title = "Knowledge Neurons in Pretrained Transformers",
    author = "Dai, Damai  and
      Dong, Li  and
      Hao, Yaru  and
      Sui, Zhifang  and
      Chang, Baobao  and
      Wei, Furu",
    editor = "Muresan, Smaranda  and
      Nakov, Preslav  and
      Villavicencio, Aline",
    booktitle = "Proceedings of the 60th Annual Meeting of the Association for Computational Linguistics (Volume 1: Long Papers)",
    month = may,
    year = "2022",
    address = "Dublin, Ireland",
    publisher = "Association for Computational Linguistics",
    url = "https://aclanthology.org/2022.acl-long.581/",
    doi = "10.18653/v1/2022.acl-long.581",
    pages = "8493--8502"
}

@inproceedings{
nanda2023progress,
title={Progress measures for grokking via mechanistic interpretability},
author={Neel Nanda and Lawrence Chan and Tom Lieberum and Jess Smith and Jacob Steinhardt},
booktitle={The Eleventh International Conference on Learning Representations },
year={2023},
url={https://openreview.net/forum?id=9XFSbDPmdW}
}

@inproceedings{
wu2025retrieval,
title={Retrieval Head Mechanistically Explains Long-Context Factuality},
author={Wenhao Wu and Yizhong Wang and Guangxuan Xiao and Hao Peng and Yao Fu},
booktitle={The Thirteenth International Conference on Learning Representations},
year={2025},
url={https://openreview.net/forum?id=EytBpUGB1Z}
}

@inproceedings{crosbie-shutova-2025-induction,
    title = "Induction Heads as an Essential Mechanism for Pattern Matching in In-context Learning",
    author = "Crosbie, Joy  and
      Shutova, Ekaterina",
    editor = "Chiruzzo, Luis  and
      Ritter, Alan  and
      Wang, Lu",
    booktitle = "Findings of the Association for Computational Linguistics: NAACL 2025",
    month = apr,
    year = "2025",
    address = "Albuquerque, New Mexico",
    publisher = "Association for Computational Linguistics",
    url = "https://aclanthology.org/2025.findings-naacl.283/",
    doi = "10.18653/v1/2025.findings-naacl.283",
    pages = "5049--5111",
    ISBN = "979-8-89176-195-7"
}

@inproceedings{huo-etal-2024-mmneuron,
    title = "{MMN}euron: Discovering Neuron-Level Domain-Specific Interpretation in Multimodal Large Language Model",
    author = "Huo, Jiahao  and
      Yan, Yibo  and
      Hu, Boren  and
      Yue, Yutao  and
      Hu, Xuming",
    editor = "Al-Onaizan, Yaser  and
      Bansal, Mohit  and
      Chen, Yun-Nung",
    booktitle = "Proceedings of the 2024 Conference on Empirical Methods in Natural Language Processing",
    month = nov,
    year = "2024",
    address = "Miami, Florida, USA",
    publisher = "Association for Computational Linguistics",
    url = "https://aclanthology.org/2024.emnlp-main.387/",
    doi = "10.18653/v1/2024.emnlp-main.387",
    pages = "6801--6816"
}

@inproceedings{dar-etal-2023-analyzing,
    title = "Analyzing Transformers in Embedding Space",
    author = "Dar, Guy  and
      Geva, Mor  and
      Gupta, Ankit  and
      Berant, Jonathan",
    editor = "Rogers, Anna  and
      Boyd-Graber, Jordan  and
      Okazaki, Naoaki",
    booktitle = "Proceedings of the 61st Annual Meeting of the Association for Computational Linguistics (Volume 1: Long Papers)",
    month = jul,
    year = "2023",
    address = "Toronto, Canada",
    publisher = "Association for Computational Linguistics",
    url = "https://aclanthology.org/2023.acl-long.893/",
    doi = "10.18653/v1/2023.acl-long.893",
    pages = "16124--16170"
}

@inproceedings{
stolfo2023a,
title={A Mechanistic Interpretation of Arithmetic Reasoning in Language Models using Causal Mediation Analysis},
author={Alessandro Stolfo and Yonatan Belinkov and Mrinmaya Sachan},
booktitle={The 2023 Conference on Empirical Methods in Natural Language Processing},
year={2023},
url={https://openreview.net/forum?id=aB3Hwh4UzP}
}

@misc{ma2026interpretabilityperformanceoptimizingretrieval,
      title={From Interpretability to Performance: Optimizing Retrieval Heads for Long-Context Language Models}, 
      author={Youmi Ma and Naoaki Okazaki},
      year={2026},
      eprint={2601.11020},
      archivePrefix={arXiv},
      primaryClass={cs.CL},
      url={https://arxiv.org/abs/2601.11020}, 
}

@inproceedings{
meng2022locating,
title={Locating and Editing Factual Associations in {GPT}},
author={Kevin Meng and David Bau and Alex J Andonian and Yonatan Belinkov},
booktitle={Advances in Neural Information Processing Systems},
editor={Alice H. Oh and Alekh Agarwal and Danielle Belgrave and Kyunghyun Cho},
year={2022},
url={https://openreview.net/forum?id=-h6WAS6eE4}
}

\onecolumn
\appendix

\begin{figure*}[t]
    \centering
    \includegraphics[width=15cm]{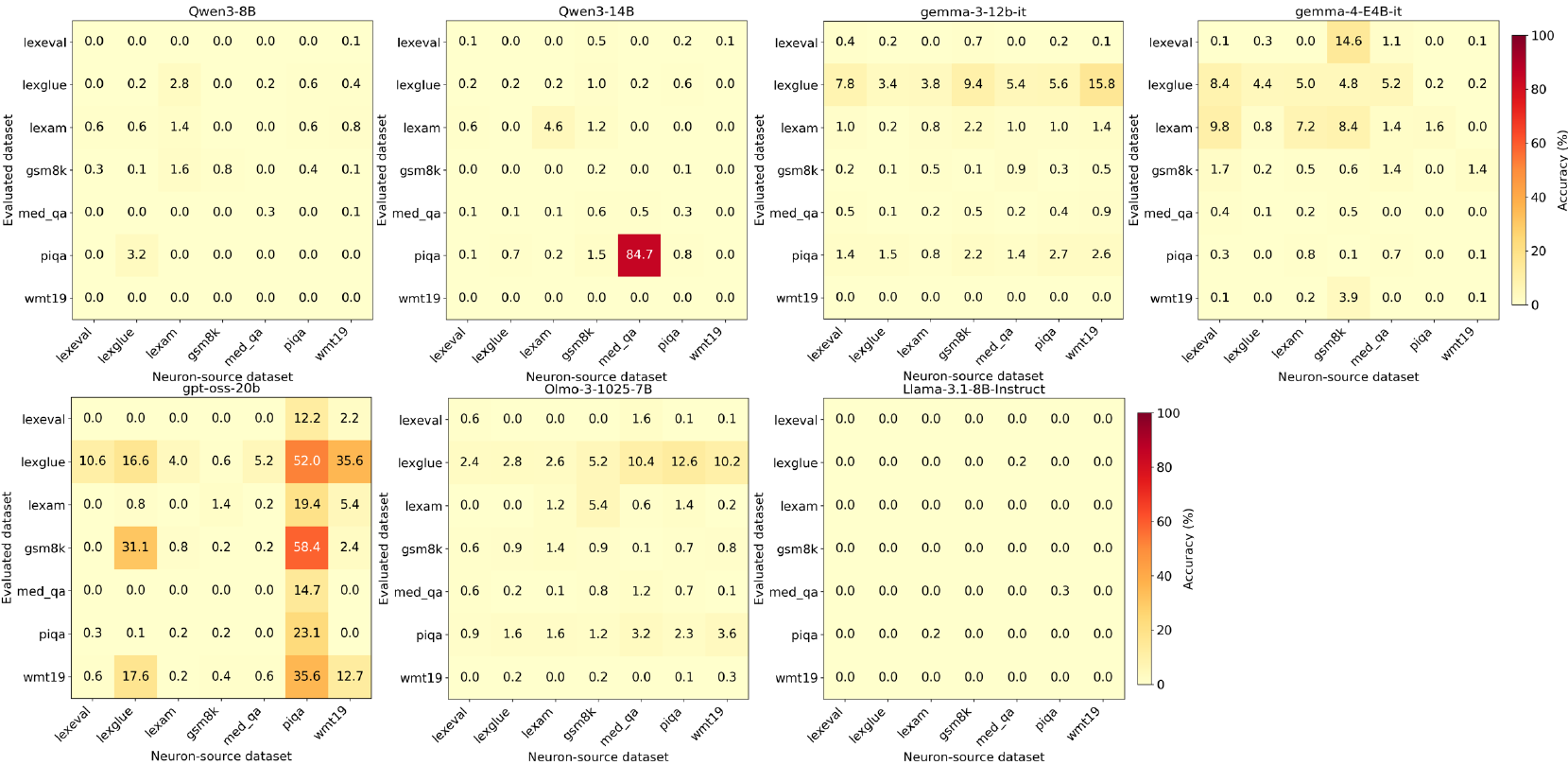}
    \caption{Neuron suppression result of top 1\% attribution score neuron.}
    \label{fig:neuron_suppress_top1per}
\end{figure*}

\begin{figure*}[t]
    \centering
    \includegraphics[width=15cm]{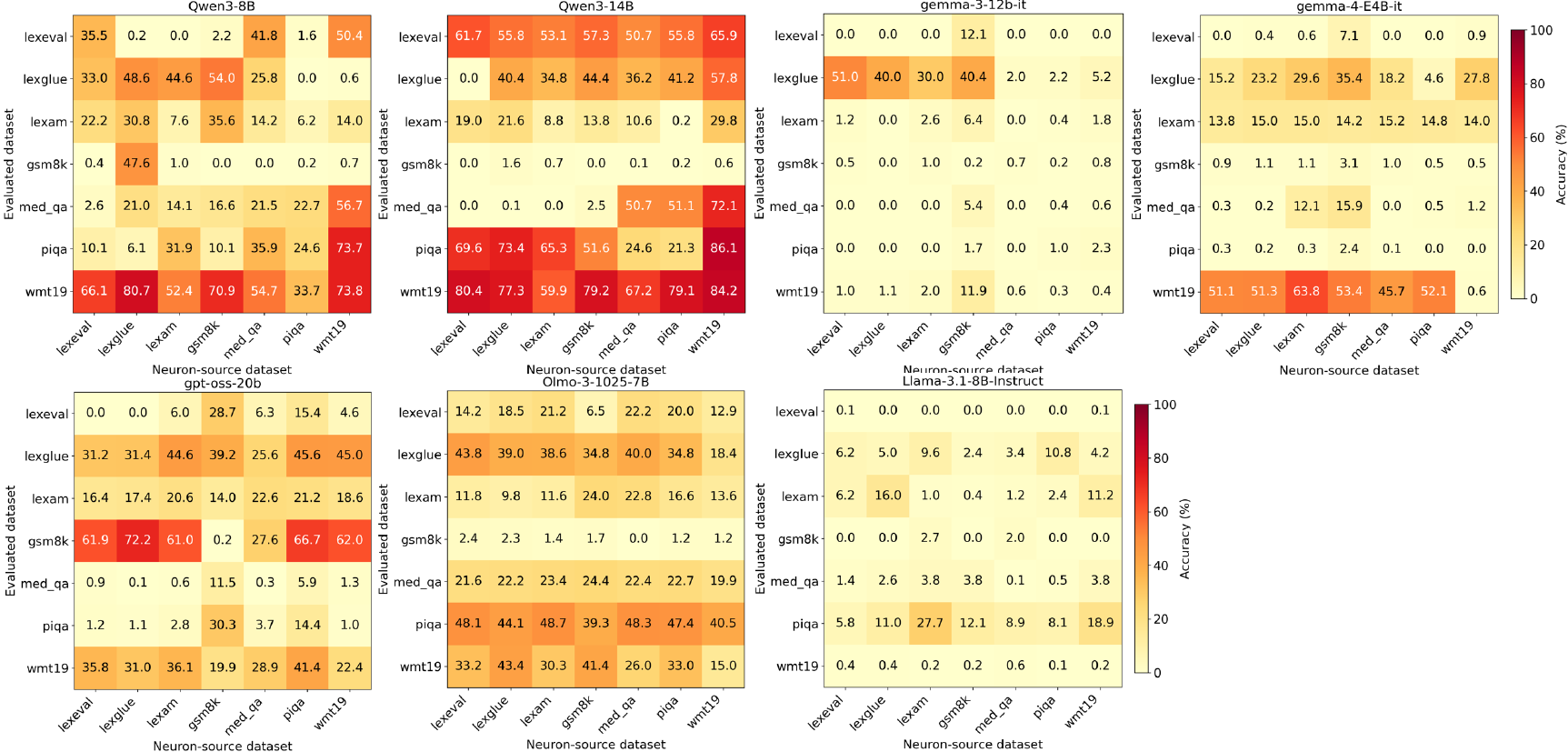}
    \caption{Neuron suppression result of top 0.3\% attribution score neuron.}
    \label{fig:neuron_suppress_top0.3per}
\end{figure*}

\begin{figure*}[t]
    \centering
    \includegraphics[width=15.5cm]{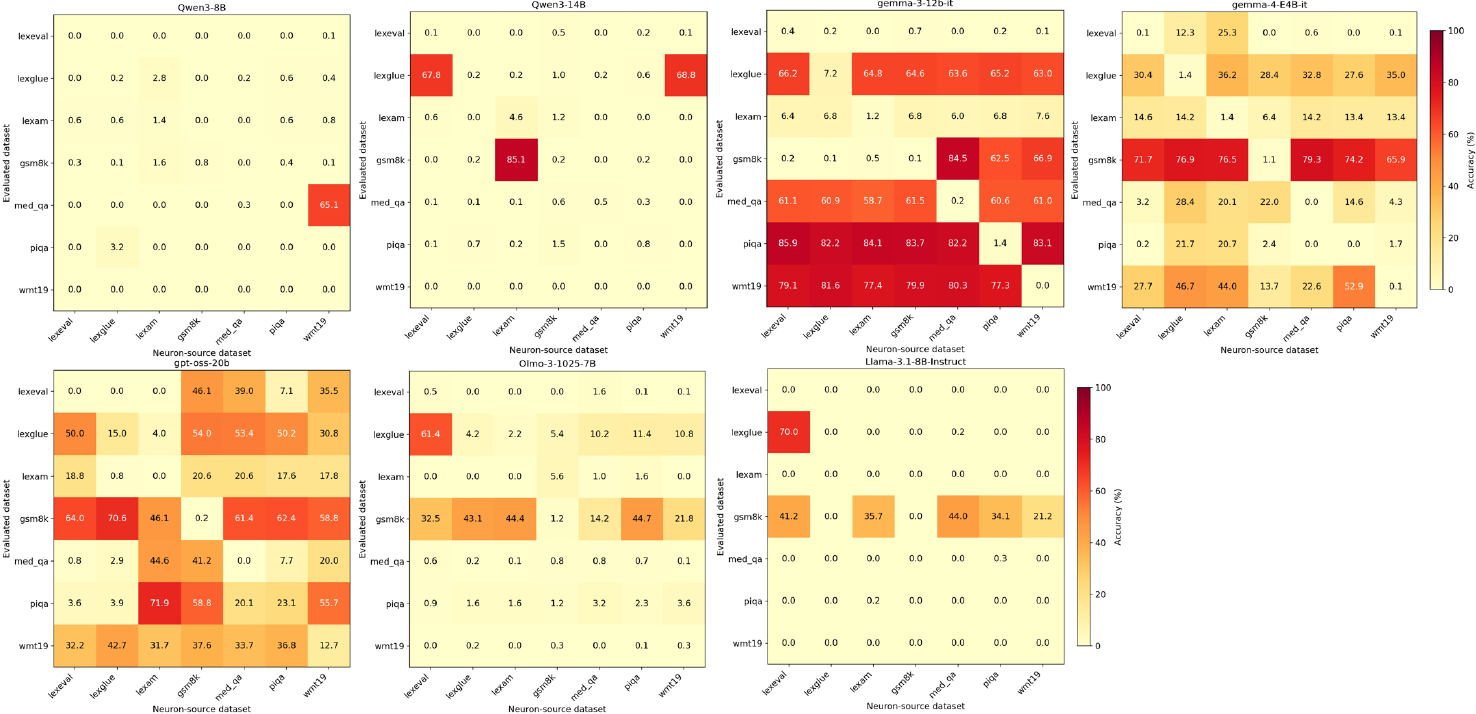}
    \caption{The result of neuron suppression in the top 0.5\% identified neurons.}
    \label{fig:neuron_suppress_0.5_all}
\end{figure*}

\begin{figure*}[t]
    \centering
    \includegraphics[width=15.5cm]{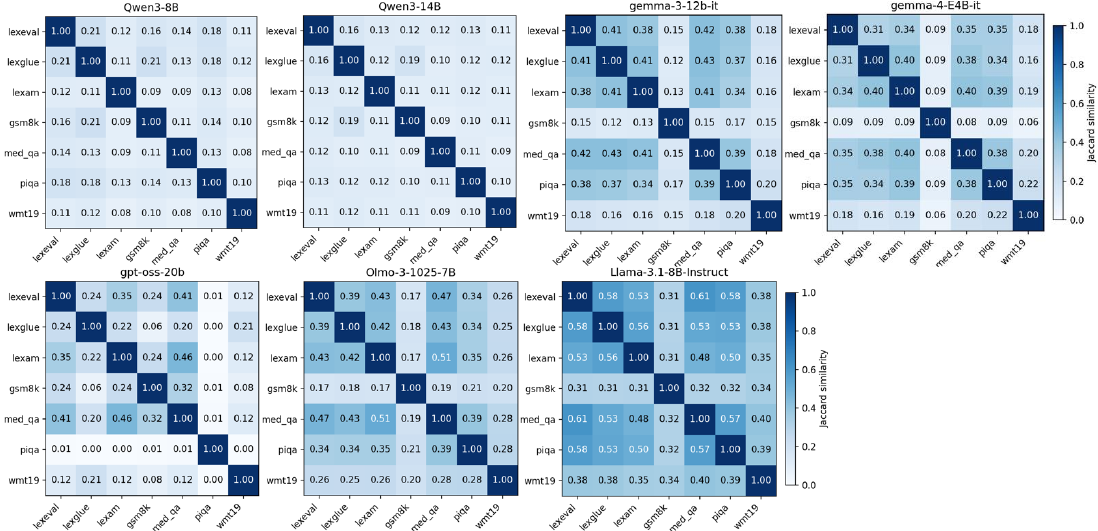}
    \caption{Neuron overlap of each model in the top 0.5\% neurons.}
    \label{fig:neuron_overlap_all}
\end{figure*}

\begin{figure*}[t]
    \centering
    \includegraphics[width=15.5cm]{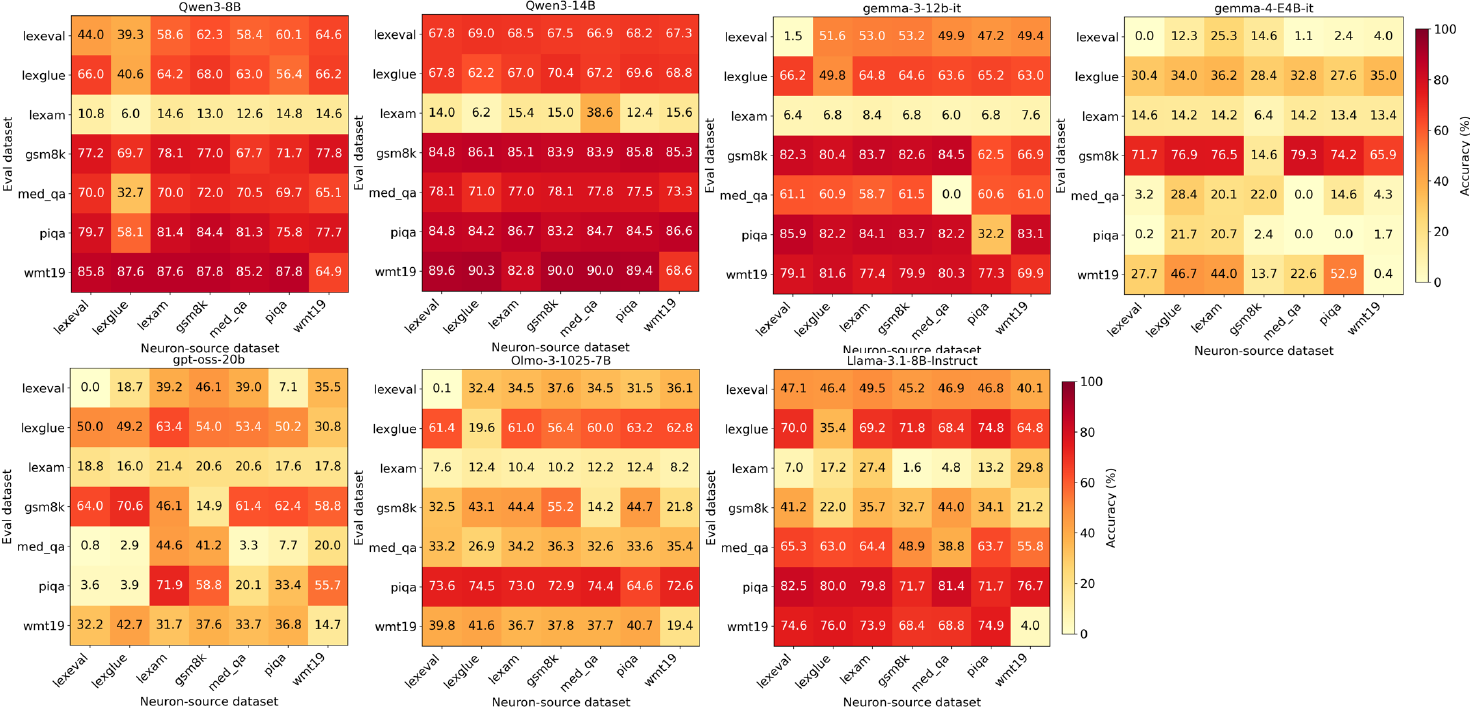}
    \caption{Neuron suppressed result when shared neurons are removed.}
    \label{fig:top_only_neurons_0.5per_all}
\end{figure*}

\begin{figure*}[t]
    \centering
    \includegraphics[width=15.5cm]{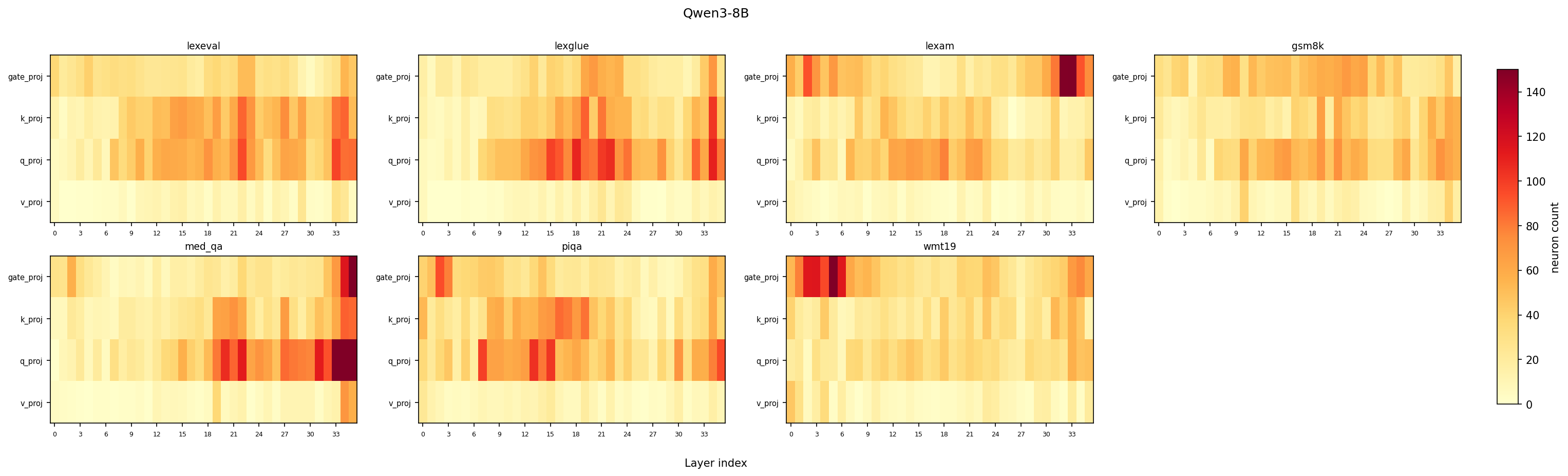}
    \caption{Neuron distribution of Qwen3-8B in top 0.5\%.}
    \label{fig:neuron_dist_Qwen3-8B}
\end{figure*}

\begin{figure*}[t]
    \centering
    \includegraphics[width=15cm]{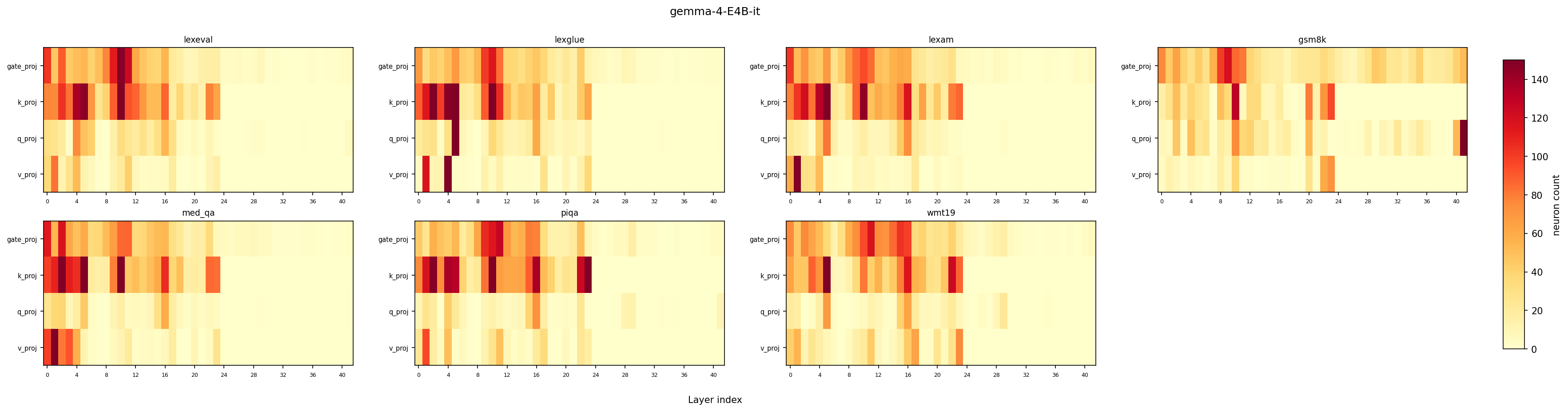}
    \caption{Neuron distribution of Gemma-4-E4B-it in top 0.5\%.}
    \label{fig:neuron_dist_gemma-4-E4B-it}
\end{figure*}

\begin{figure*}[t]
    \centering
    \includegraphics[width=15cm]{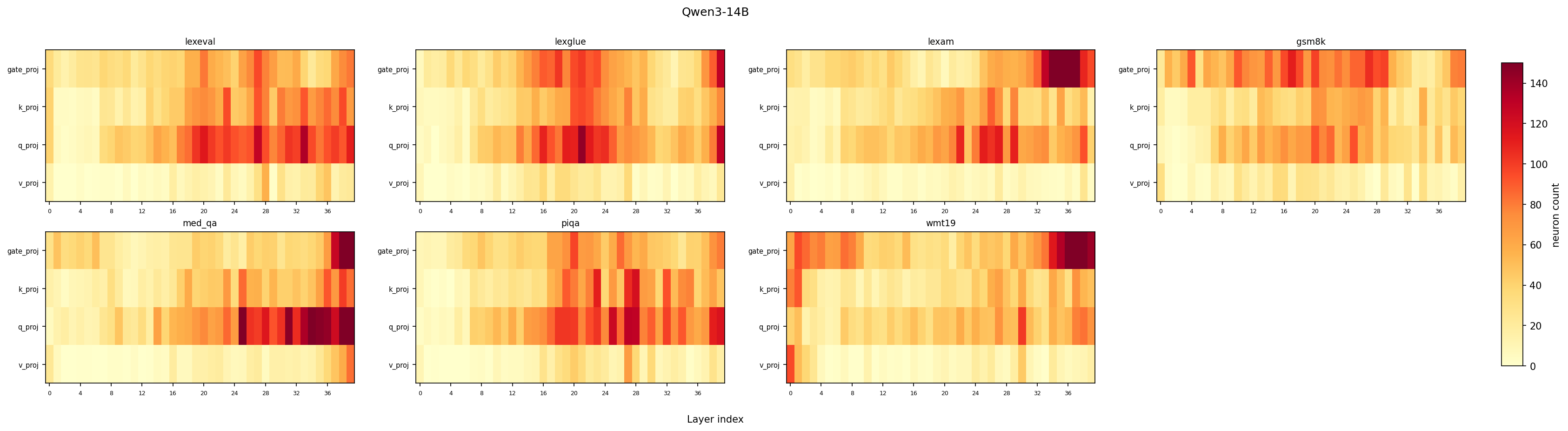}
    \caption{Neuron distribution of Qwen3-14B in top 0.5\%.}
    \label{fig:neuron_dist_Qwen3-14B}
\end{figure*}

\begin{figure*}[t]
    \centering
    \includegraphics[width=15cm]{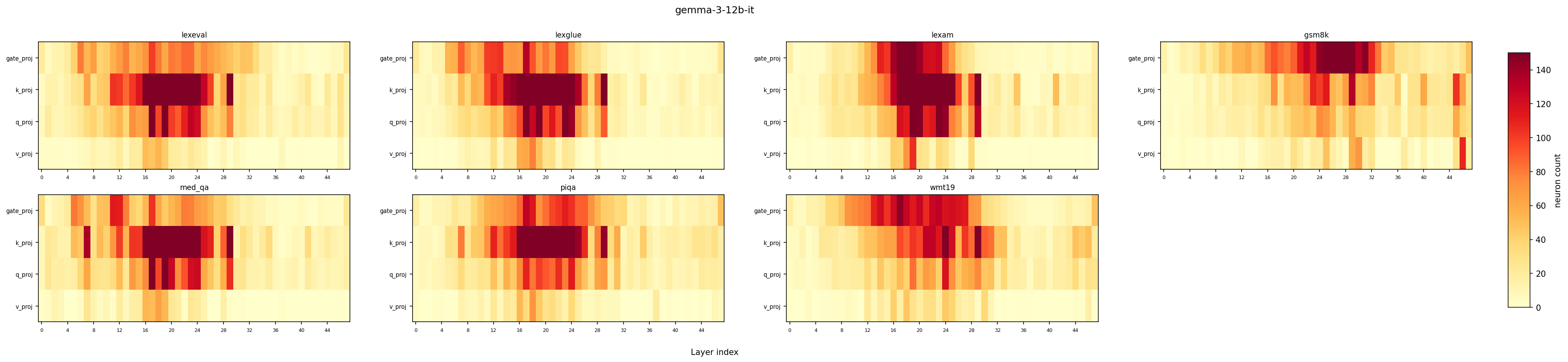}
    \caption{Neuron distribution of  Gemma-3-12B-it in top 0.5\%.}
    \label{fig:neuron_dist_gemma-3-12b-it}
\end{figure*}

\begin{figure*}[t]
    \centering
    \includegraphics[width=15cm]{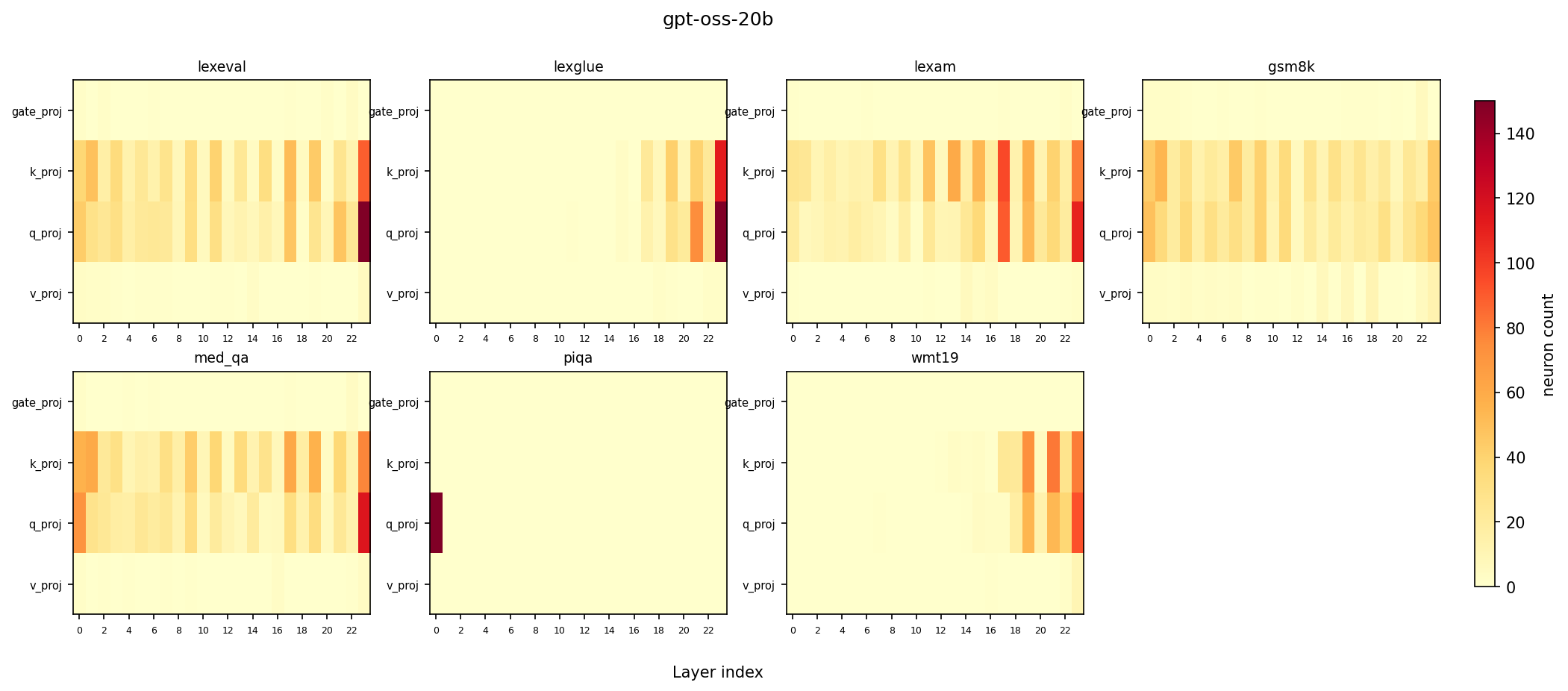}
    \caption{Neuron distribution of GPT-OSS-20B in top 0.5\%.}
    \label{fig:neuron_dist_gpt-oss-20b}
\end{figure*}

\begin{figure*}[t]
    \centering
    \includegraphics[width=15cm]{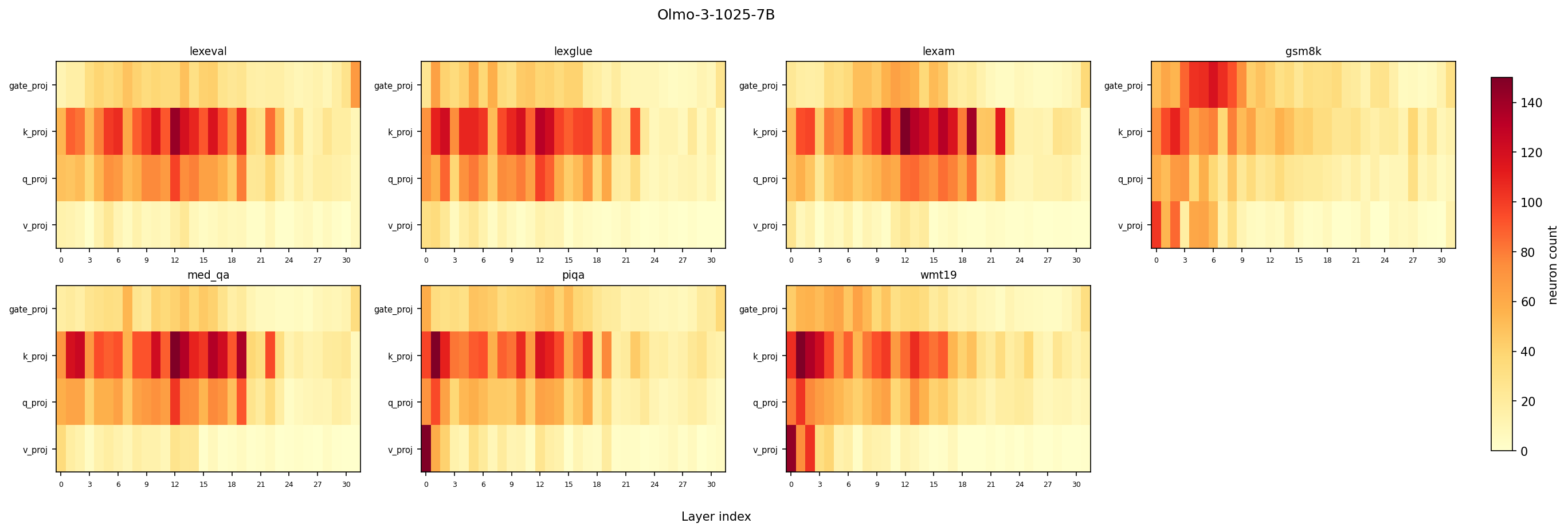}
    \caption{Neuron distribution of Olmo-3-1025-7B in top 0.5\%.}
    \label{fig:neuron_dist_Olmo-3-1025-7B}
\end{figure*}

\begin{figure*}[t]
    \centering
    \includegraphics[width=15cm]{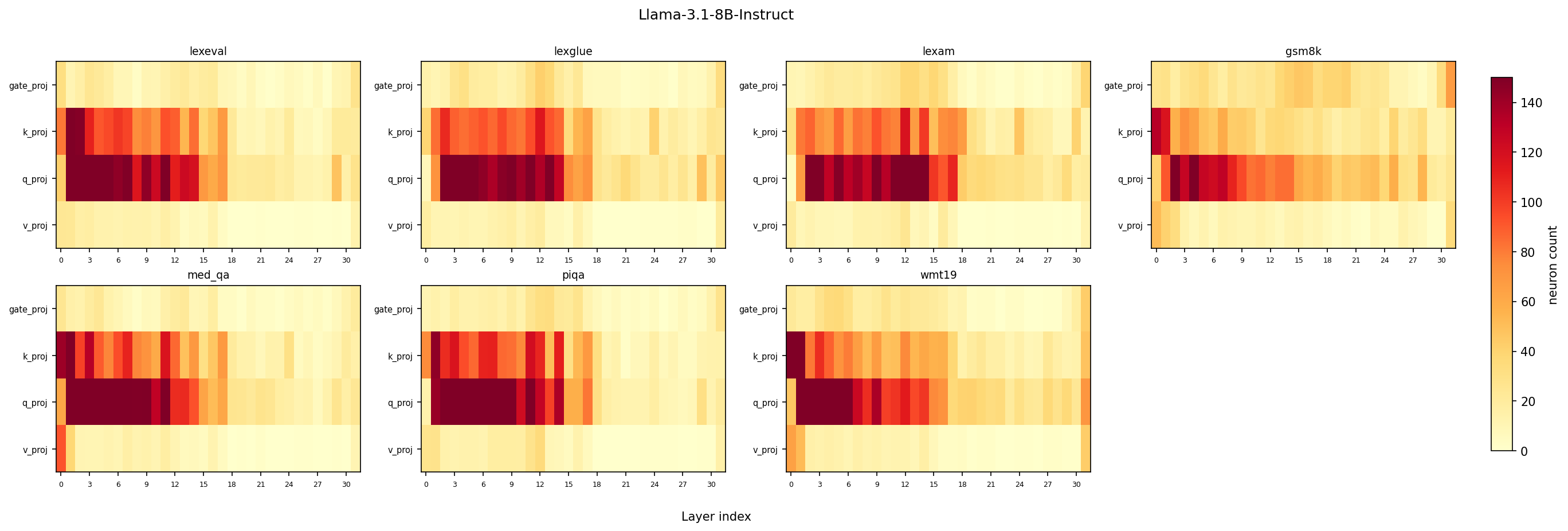}
    \caption{Neuron distribution of Llama-3.1-8B-Instruct in top 0.5\%.}
    \label{fig:neuron_dist_Llama-3.1-8B-Instruct}
\end{figure*}


\begin{figure*}[t]
    \centering
    \includegraphics[width=15cm]{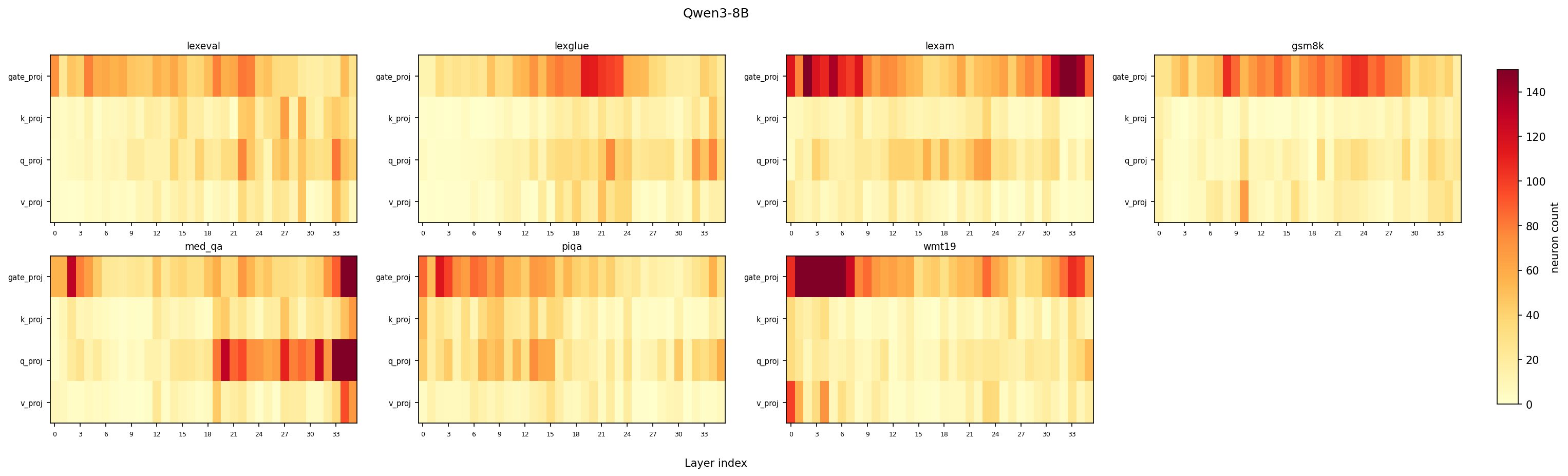}
    \caption{Neuron Distribution after shared neuron removal of Qwen3-8B.}
    \label{fig:neuron_dist_remove_Qwen3-8B}
\end{figure*}

\begin{figure*}[t]
    \centering
    \includegraphics[width=15cm]{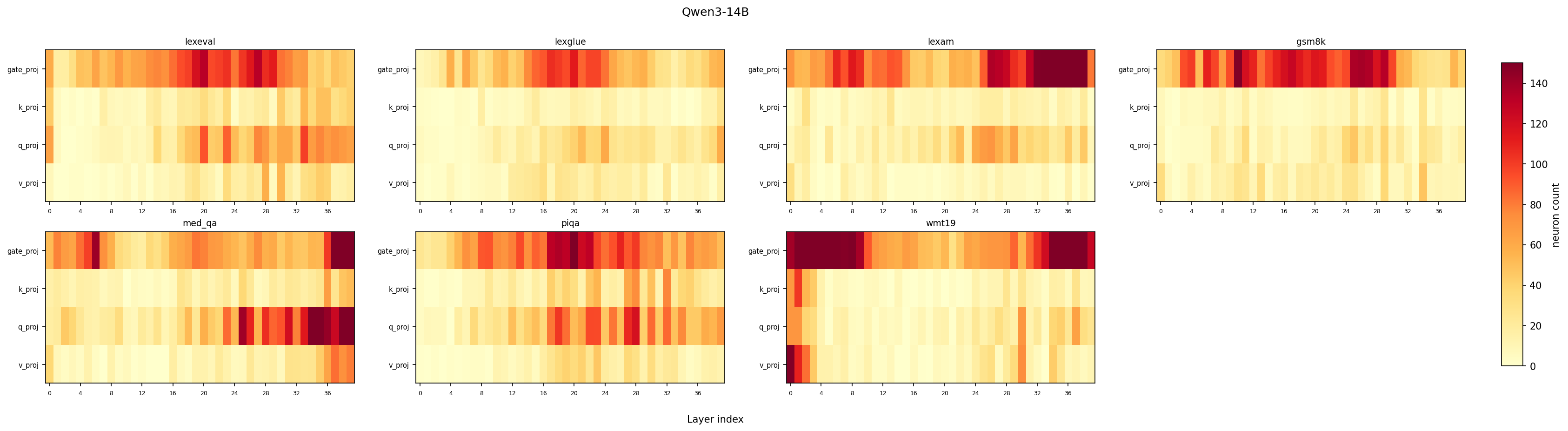}
    \caption{Neuron Distribution after shared neuron removal of Qwen3-14B.}
    \label{fig:neuron_dist_remove_Qwen3-14B}
\end{figure*}

\begin{figure*}[t]
    \centering
    \includegraphics[width=15cm]{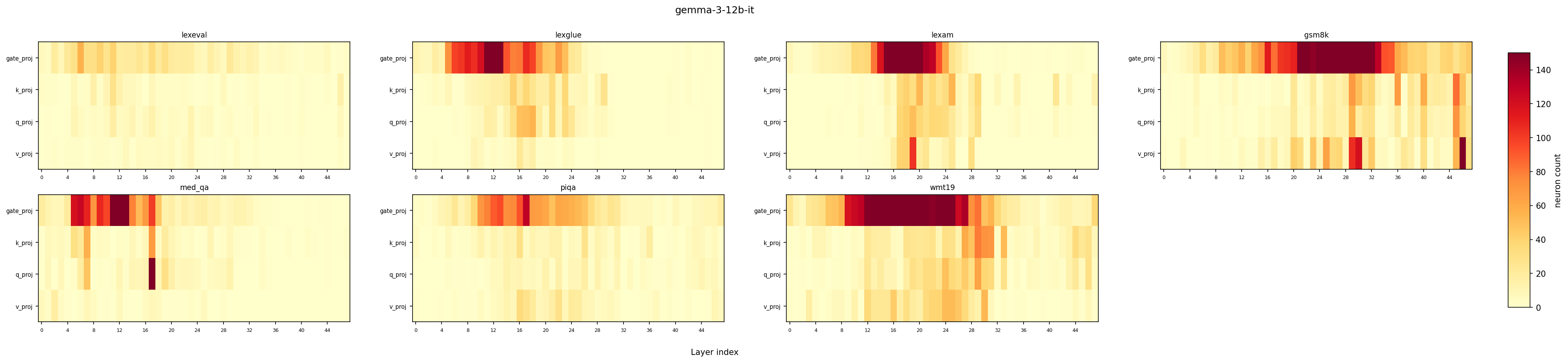}
    \caption{Neuron Distribution after shared neuron removal of  Gemma-3-12B-it.}
    \label{fig:neuron_dist_remove_gemma-3-12b-it}
\end{figure*}

\begin{figure*}[t]
    \centering
    \includegraphics[width=15cm]{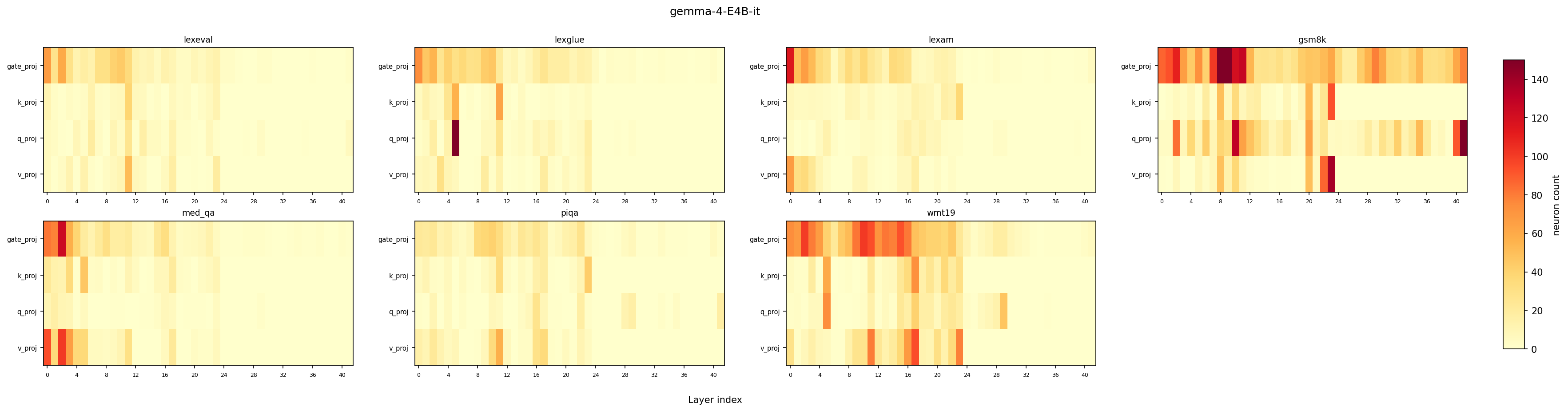}
    \caption{Neuron Distribution after shared neuron removal of Gemma-4-E4B-it.}
    \label{fig:neuron_dist_remove_gemma-4-E4B-it}
\end{figure*}

\begin{figure*}[t]
    \centering
    \includegraphics[width=15cm]{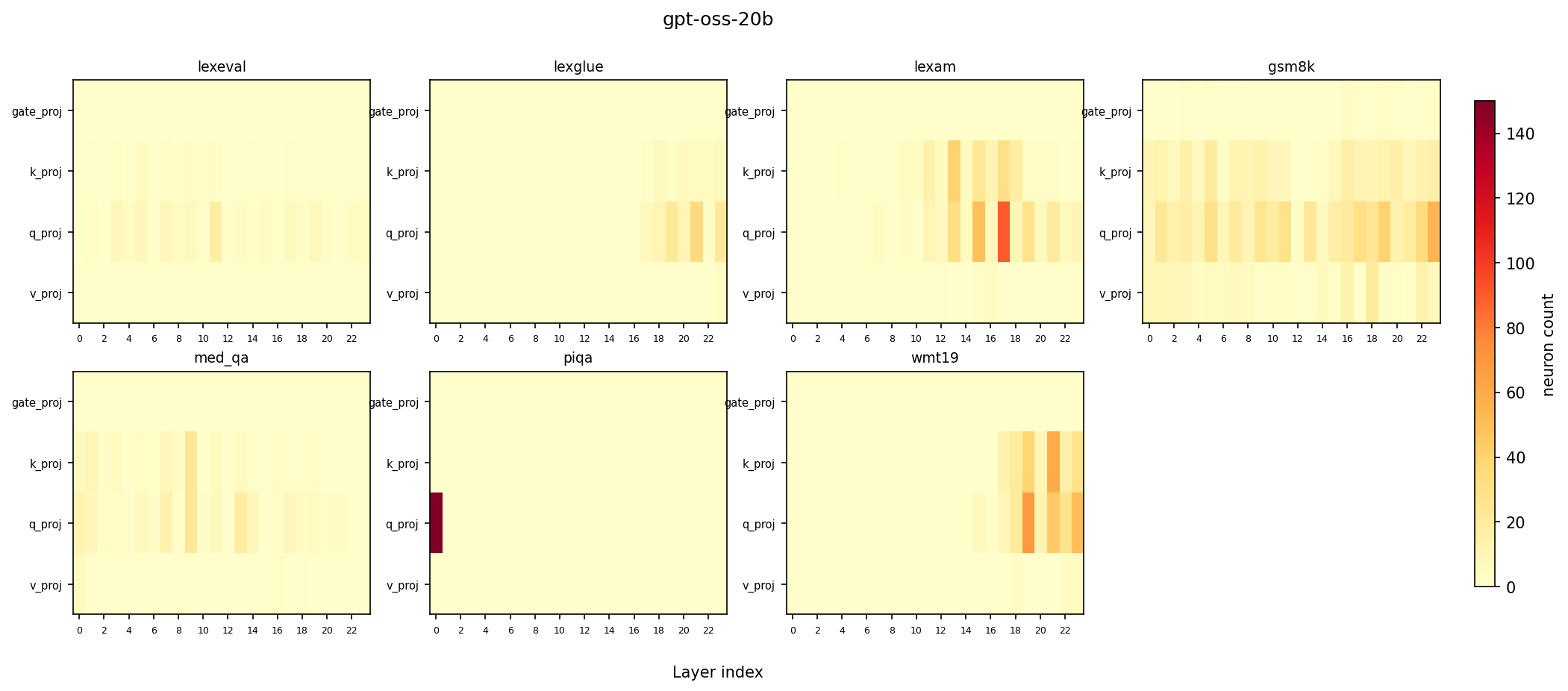}
    \caption{Neuron Distribution after shared neuron removal of GPT-OSS-20B.}
    \label{fig:neuron_dist_remove_gpt-oss-20b}
\end{figure*}

\begin{figure*}[t]
    \centering
    \includegraphics[width=15cm]{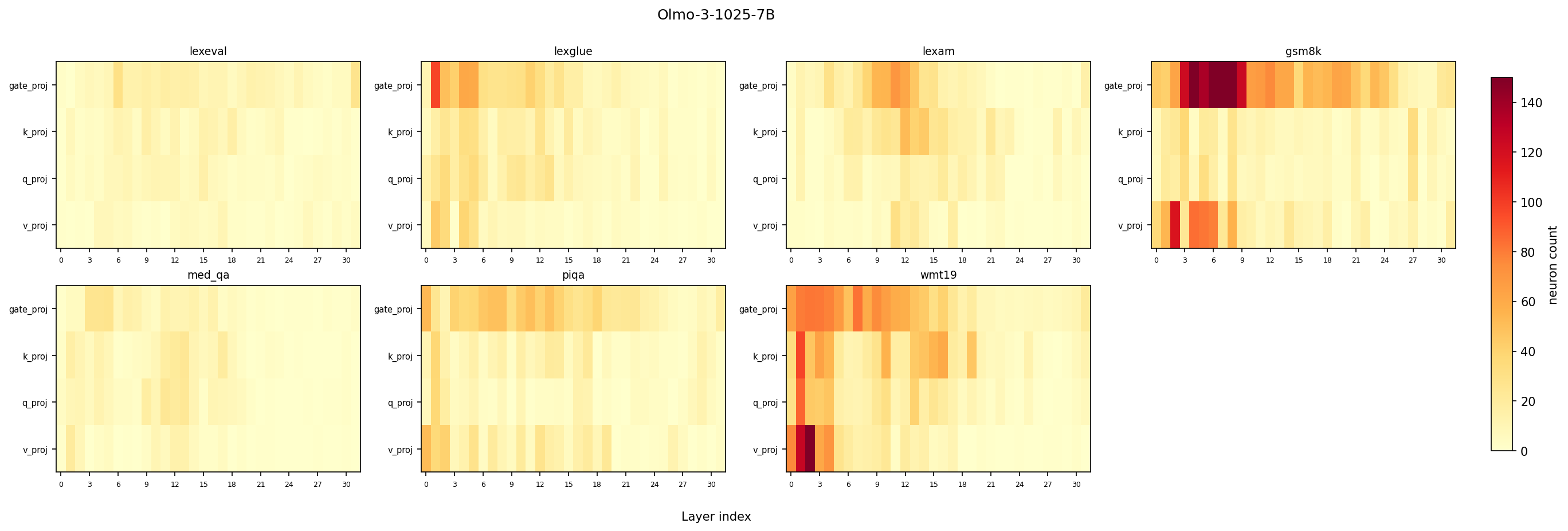}
    \caption{Neuron Distribution after shared neuron removal of Olmo-3-1025-7B.}
    \label{fig:neuron_dist_remove_Olmo-3-1025-7B}
\end{figure*}

\begin{figure*}[t]
    \centering
    \includegraphics[width=15cm]{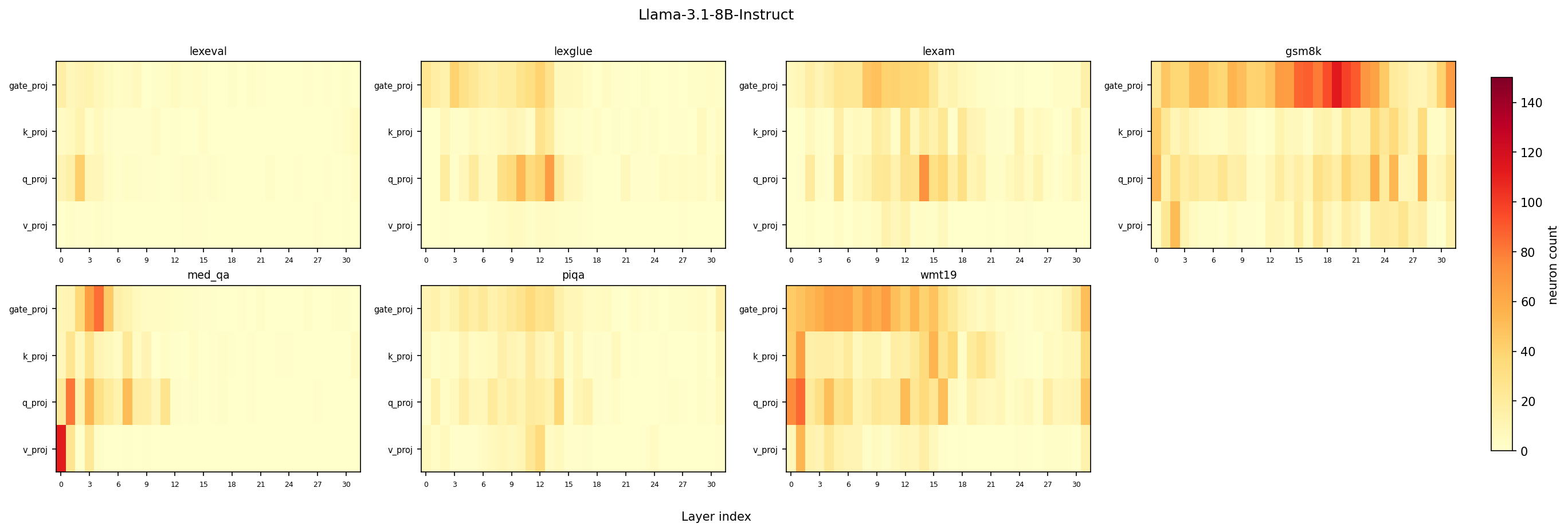}
    \caption{Neuron Distribution after shared neuron removal of Llama-3.1-8B-Instruct.}
    \label{fig:neuron_dist_remove_Llama-3.1-8B-Instruct}
\end{figure*}

\section{Top 1\% and top 3\% neuron suppression}
\label{sec:top1and0.3_neuron_suppression_appendix}

Figures~\ref{fig:neuron_suppress_top1per} and ~\ref{fig:neuron_suppress_top0.3per} present the results of suppressing the top 1\% and top 0.3\% of neurons.

In the top 1\% neuron suppression experiment, performance drops to nearly zero across all tasks, likely because the method selects too many neurons, capturing not only task-specific neurons but also a large number of more generally important neurons.

In the results obtained by suppressing the top 0.3\% of neurons, there are cases in Qwen3-8B, Qwen3-14B, Gemma-4-E4B-it, GPT-OSS-20B, and Olmo-3-1025-7B where task-specific neurons appear to have been successfully identified. Specifically, for LEXam and PIQA in Qwen3-8B, PIQA in Qwen3-14B, WMT19 in Gemma-4-E4B-it, GSM8K and WMT19 in GPT-OSS-20B, and WMT19 in Olmo-3-1025-7B, suppressing the selected neurons reduces model performance only on the corresponding task, while performance on the other tasks remains comparatively high.

\section{Top 0.5\% scored neuron suppression}
\label{sec:top0.5_scored_neuron_suppression_appendix}

Figure~\ref{fig:neuron_suppress_0.5_all} and Figure~\ref{fig:neuron_overlap_all} show the full result of neuron suppression in identified neurons and identified neuron overlap between each tasks in models. 

\section{Influential neuron distribution for top 0.5\% scored neurons}
\label{sec:important_neuron_distribution_appendix}

Figure~\ref{fig:neuron_dist_Qwen3-8B}, Figure~\ref{fig:neuron_dist_Qwen3-14B}, Figure~\ref{fig:neuron_dist_gemma-3-12b-it}, 
Figure~\ref{fig:neuron_dist_gemma-4-E4B-it}, Figure~\ref{fig:neuron_dist_gpt-oss-20b}, Figure~\ref{fig:neuron_dist_Olmo-3-1025-7B}, and Figure~\ref{fig:neuron_dist_Llama-3.1-8B-Instruct} shows the important neuron distribution in top 0.5\% of identified neurons in the experiment of Section~\ref{sec:discussion}.
In Llama-3.1-8B-Instruct and GPT-OSS-20B, important neurons are gathering on the K and Q outputs of the Attention.

Given that the existing study of important neuron distribution analyzed a short and general text understanding task, 
This result suggests that our method captures mechanisms shared across LLMs.

\section{Influential neuron distribution after the shared neuron removal}
\label{sec:important_neuron_shared_neuron_removal_appendix}

The complete result of shared neuron removal is shown in Figure~\ref{fig:top_only_neurons_0.5per_all}.
Figure~\ref{fig:neuron_dist_remove_Qwen3-8B}, Figure~\ref{fig:neuron_dist_remove_Qwen3-14B}, Figure~\ref{fig:neuron_dist_remove_gemma-3-12b-it}, 
Figure~\ref{fig:neuron_dist_remove_gemma-4-E4B-it}, Figure~\ref{fig:neuron_dist_remove_gpt-oss-20b}, Figure~\ref{fig:neuron_dist_remove_Olmo-3-1025-7B}, and Figure~\ref{fig:neuron_dist_remove_Llama-3.1-8B-Instruct} shows the result after shared neuron removal in Section~\ref{sec:important_neuron_shared_neuron_removal}.

In  Gemma-3-12B-it and Gemma-4-E4B-it, the prior hypothesis that important neurons are concentrated in the middle MLP layers is confirmed more clearly across nearly all tasks. A similar tendency toward concentration in the middle MLP layers is also observed for LEXam in Olmo-3-1025-7B, and for LEXam, LexGLUE, MedQA, and PIQA in Llama-3.1-8B-Instruct. In Qwen3-8B and Qwen3-14B, unlike the case where the top 5\% of neurons are selected solely by neuron attribution score, important neurons appear to be distributed broadly across all MLP layers. In GPT-OSS-20B, as in the analysis based on the top 5\% of neurons by neuron attribution score, the K and Q components of the attention mechanism remain particularly important even after removing neurons that are shared across the seven tasks.

\begin{figure*}[t]
    \centering
    \includegraphics[width=15cm]{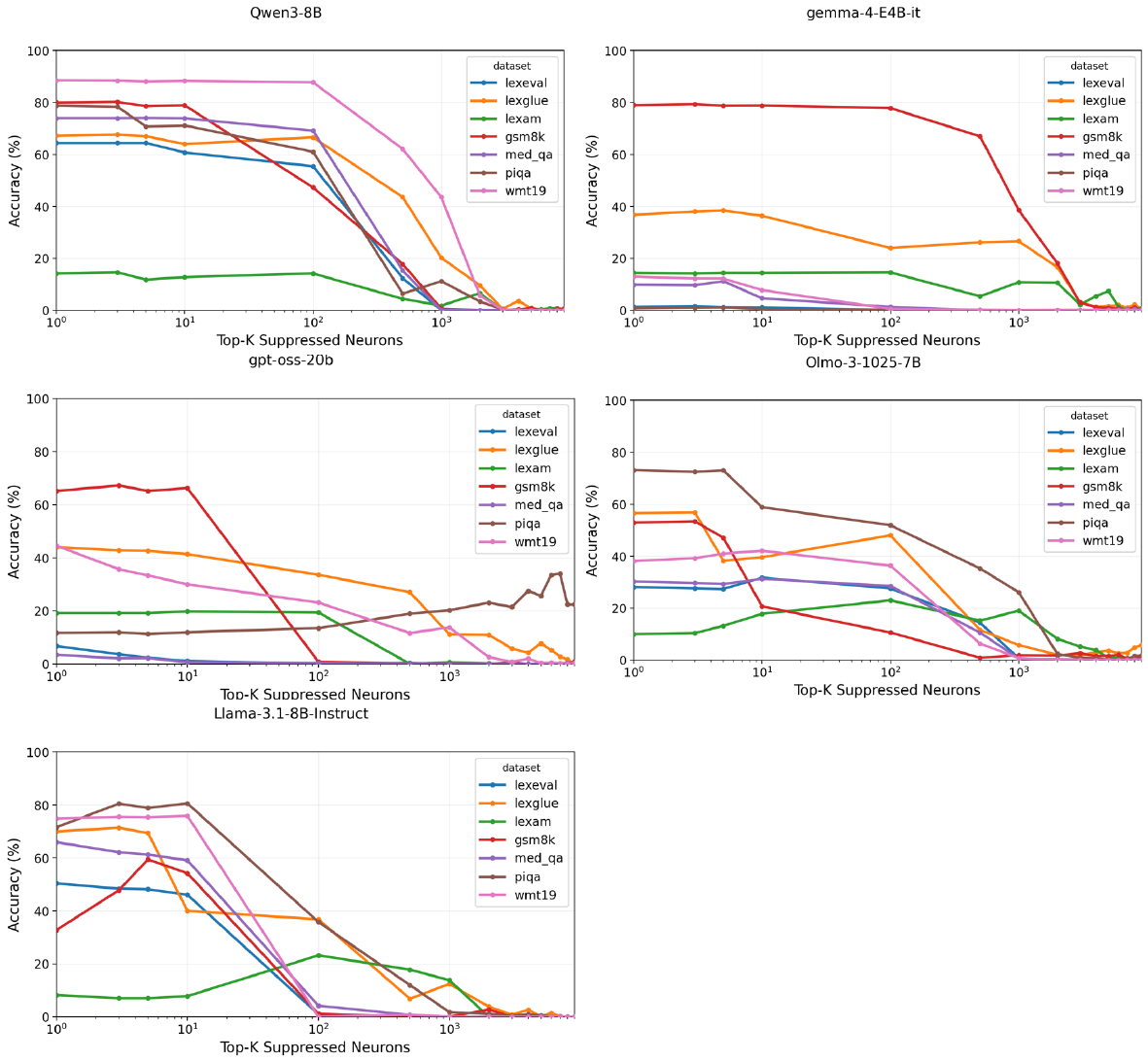}
    \caption{Top K neuron suppress sorted by neuron attribution score in each dataset.}
    \label{fig:topk_suppression_appendix}
\end{figure*}

\section{Score change by top k neuron suppression}
\label{sec:score_change_by_topk_neuron_suppression_appendix}

Figure~\ref{fig:topk_suppression_appendix} shows a comprehensive analysis of the relation and the number of suppressed neurons and task performance for the identified neurons in each task.
Qwen3-8B and Qwen3-14B exhibit a sharp decline in accuracy when suppressing between the top 100 and top 1,000 neurons. In  Gemma-3-12B-it and Llama-3.1-8B-Instruct, task accuracy begins to decline between the top 10 and top 100 neurons. By contrast, in Gemma-4-E4B-it and Olmo-3-1025-7B, the decrease is more gradual, with substantial accuracy degradation emerging around the top 1,000 neurons. In GPT-OSS-20B, GSM8K accuracy declines between the top 10 and top 100 neurons, whereas performance on the other tasks remains relatively stable, showing only limited degradation.

\section{Experimental settings and AI usage}
In our experiments, we used NVIDIA H100 and H200 GPUs.
We used AI tools for translation of analyzed datasets and model outputs, translation of our manuscript, and proofreading.

\end{document}